\documentclass[sigconf]{acmart}

\usepackage{booktabs} 

\setcopyright{acmcopyright}

\copyrightyear{2018} 
\acmYear{2018} 
\setcopyright{acmcopyright}
\acmConference[GECCO '18]{Genetic and Evolutionary Computation Conference}{July 15--19, 2018}{Kyoto, Japan}
\acmBooktitle{GECCO '18: Genetic and Evolutionary Computation Conference, July 15--19, 2018, Kyoto, Japan}
\acmPrice{15.00}

\acmDOI{10.1145/3205455.3205648}

\acmISBN{978-1-4503-5618-3/18/07}

\editor{Jennifer B. Sartor}
\editor{Theo D'Hondt}
\editor{Wolfgang De Meuter}

\usepackage[linesnumbered,vlined,ruled,boxed,commentsnumbered]{algorithm2e}
  \usepackage{algpseudocode}
  \usepackage{bm} 

\newcommand{\tref}[1]{Table \ref{#1}}
\newcommand{\sref}[1]{Section \ref{#1}}
\newcommand{\eref}[1]{Eq. \ref{#1}}
\newcommand{\aref}[1]{Alg. \ref{#1}}
\usepackage{cleveref}

\usepackage{float}
\usepackage{subfigure} 

\usepackage{threeparttable}
\usepackage{multirow}
\usepackage{bbm}

\usepackage{color, colortbl}

\usepackage{verbatim}

\begin{document}

\title{MOEA/D with Uniformly Randomly Adaptive Weights}



\author{Lucas R. C. de Farias}
\affiliation{%
  \institution{Universidade Federal de Pernambuco}
  \city{Recife} 
  \country{Brazil}
}
\email{lrcf@cin.ufpe.br}

\author{Pedro H. M. Braga}
\affiliation{%
  \institution{Universidade Federal de Pernambuco}
  \city{Recife} 
  \country{Brazil}
}
\email{phmb4@cin.ufpe.br}

\author{Hansenclever F. Bassani}
\affiliation{%
  \institution{Universidade Federal de Pernambuco}
  \city{Recife} 
  \country{Brazil}}
\email{hfb@cin.ufpe.br}

\author{Aluizio F. R. Ara\'ujo}
\affiliation{%
  \institution{Universidade Federal de Pernambuco}
  \city{Recife}
  \country{Brazil}
}
\email{aluizioa@cin.ufpe.br}


\begin{abstract}
When working with decomposition-based algorithms, an appropriate set of weights might improve quality of the final solution. A set of uniformly distributed weights usually leads to well-distributed solutions on a Pareto front. However, there are two main difficulties with this approach. Firstly, it may fail depending on the problem geometry. Secondly, the population size becomes not flexible as the number of objectives increases. In this paper, we propose the MOEA/D with Uniformly Randomly Adaptive Weights (MOEA/D-URAW) which uses the Uniformly Randomly method as an approach to subproblems generation, allowing a flexible population size even when working with many objective problems. During the evolutionary process, MOEA/D-URAW adds and removes subproblems as a function of the sparsity level of the population. Moreover, instead of requiring assumptions about the Pareto front shape, our method adapts its weights to the shape of the problem during the evolutionary process. Experimental results using WFG41-48 problem classes, with different Pareto front shapes, shows that the present method presents better or equal results in 77.5\% of the problems evaluated from 2 to 6 objectives when compared with state-of-the-art methods in the literature.
\end{abstract}

%
%

\ccsdesc[500]{Theory of computation~Evolutionary algorithms}

\keywords{Many-objective optimization, Multi-objective optimization, Evolution strategies, Decomposition methods}

\maketitle

\section{Introduction}

	Decomposition-based evolutionary multiobjective optimization (EMO) algorithms decomposes a multiobjective optimization problem (MOOP) \cite{deb2001multi} into a number of single-objective optimization problems using a set of weight vectors \cite{zhang2007moea}. Each subproblem or weight vector is associated with a solution in the population, and the diversity of the evolutionary population is controlled explicitly by a set of weight vectors \cite{li2017weights}. Thus, an appropriate set of weights can increase the quality of the final solution.

A set of uniformly distributed weights usually leads to well-distributed solutions on a Pareto front (PF). However, there are two main difficulties with this approach. Firstly, it may fail depending on the problem geometry. When dealing with a complex Pareto Front (e.g., disconnected, degenerate, badly-scaled or inverted simplex-like), the final solution set can present results do not meet the expectations \cite{qi2014moea,li2017weights,wang2015preference}. Secondly, the population size becomes not flexible as the objectives number grows, because it behaves non-linearly, that is, when working with Many-objectives Optimization Problem (MaOP), the computational cost significantly increases \cite{wang2015preference}. 

To handle such limitations, we propose a method called MOEA/D with Uniformly Randomly Adaptive Weights (MOEA/D-URAW) that adapts the subproblems based on the sparsity level of the population. This approach allows a flexible population size for the MaOPs. We compared our method with other MOEA/D's variants that use fixed weight vectors for a set of test problems with different PF geometries. Experimental results show that the MOEA/D-URAW presents better or equal results in 77,5\% of the problems evaluated from 2 to 6 objectives.

The rest of this paper is organized as follows: \sref{sec:back-mot} discusses the background knowledge used in this work. \sref{sec:prop-method} introduces the proposed method MOEA/D-URAW. Thereafter, \sref{sec:setup} and \sref{sec:studies} present the experimental setup and results analysis respectively. Finally, the conclusions are presented in \sref{sec:conclusions}.

\section{Background}
\label{sec:back-mot}

	This section presents briefly the Tchebycheff (TCH) decomposition approach and the fixed weights generation methods used in the experiments of this paper. Such a decomposition is the basis for the rest of the paper.

\subsection{Tchebycheff Decomposition Approach}

In this approach, the scalar optimization problems are defined as:
\begin{equation}
\begin{split}
\label{eq:tchebycheff}
\text{minimize } & g^{TCH}(\boldsymbol{x} \vert \boldsymbol{\lambda} , \boldsymbol{z^*}) = \text{max}_{1 \leq j \leq m} (\lambda_j \vert f_j(\boldsymbol{x}) - z_j^* \vert),\\
&\text{subject to } \boldsymbol{x} \in \Omega
\end{split}
\end{equation}
where the utopian objective vector \boldsymbol{$z^*$} = ($z_1^*$ ... $z_m^*$)$^T$ is the reference point, i. e. $z_j^*$ = min $\{ f_j(\boldsymbol{x})\vert \boldsymbol{x} \in \Omega \}$, for each $j = 1,...,m$. The \textit{m}-dimensional weight vector is defined as \boldsymbol{$\lambda$} = ($\lambda_1$ ... $\lambda_m$)$^T$, $\sum_{i}^{m} \lambda_i = 1$ and $\lambda_i \geq 0$, for all $i \in 1,...,m$ \cite{zhang2007moea}. By altering weight vectors, different Pareto-optimal solutions can be obtained by the TCH approach \cite{wu2017adaptive}

\subsection{Fixed Weights Generation Methods}
\label{sec:w_generation}

Three different methods to generate a set of fixed weight vectors are described in the following \sref{sec:dd}, \sref{sec:rand} and \sref{sec:adaw}. \sref {sec:WS} presents the WS-tranformation used in the three described methods.

\subsubsection{Das and Dennis (DD)}
\label{sec:dd}
Most decomposition-based EMO algorithms use the method proposed by Das and Dennis \cite{das1998normal} to systematically generate a set of fixed weight vectors uniformly distributed over a unit simplex. Let \textit{H} be the number of divisions of each axis, totally \textit{N} = $\binom{H + m - 1}{m - 1}$ weight vectors can be generated using this approach. Since \textit{H} should be no smaller than \textit{m} to prevent intermediate points being created. The number of generated weight vectors can be very high for more than three objectives.

\subsubsection{Uniform Randomly (UR)}
\label{sec:rand}
A set of \textit{N} weight vectors \textit{W} are generated as follows. Firstly, 5000 weight vectors are uniformly random generated for forming the set $W_1$. \textit{W} is initialized as the set containing all the weight vector (1 0 ... 0 0), (0 1 ... 0 0), ... , (0 0 ... 0 1). Secondly, find the weight vector in $W_1$ with the largest distance to \textit{W}, add it to \textit{W}, and remove it from $W_1$. Then, if the size of \textit{W} is \textit{N}, stop and return \textit{W}. Otherwise, go to the second item and repeat the process \cite{zhang2009performance}.

\subsubsection{TCH Scalarizing Function (TSF)}
\label{sec:adaw}
Given a reference point, \boldsymbol{$z^*$}, the optimal weight vector to a solution with respect to Tchebycheff scalarizing function can be easily generated. This is a frequent approach in the weight vector adaptation \cite{qi2014moea,li2017weights,gu2012multiobjective}.
Formally, let \boldsymbol{$z^*$} = ($z_1^*$ ... $z_m^*$)$^T$ be the reference point and \boldsymbol{$w$} = ($\lambda_1$ $\lambda_2$ ...  $\lambda_m$)$^T$ be the optimal weight vector to a solution $\boldsymbol{q}$. Then \eref{eq:adaw-1} holds:

\begin{equation}
\begin{split}
\label{eq:adaw-1}
\frac{f_1(\boldsymbol{q}) - z_1^*}{\lambda_1} = \frac{f_2(\boldsymbol{q}) - z_2^*}{\lambda_2} = \cdot\cdot\cdot = \frac{f_m(\boldsymbol{q}) - z_m^*}{\lambda_m}
\end{split}
\end{equation}

Since $\lambda_1$ + $\lambda_2$ + ... + $\lambda_m$ = 1, we have \eref{eq:adaw-2}.
\begin{equation}
\begin{split}
\label{eq:adaw-2}
\boldsymbol{w} = (\lambda_1, \cdot\cdot\cdot, \lambda_m) = (\frac{f_1(\boldsymbol{q}) - z_1^*}{\sum_{i=1}^{m}{f_i(\boldsymbol{q}) - z_i^*}}, \cdot\cdot\cdot, \frac{f_m(\boldsymbol{q}) - z_m^*}{\sum_{i=1}^{m}{f_i(\boldsymbol{q}) - z_i^*}})
\end{split}
\end{equation}

\subsubsection{WS-Transformation}
\label{sec:WS}
It maps the weight vector of a scalar subproblem to its solution mapping vector \cite{qi2014moea}. If \boldsymbol{$\lambda$} is a weight vector, \boldsymbol{$\lambda$} = ($\lambda_1$ ... $\lambda_m$)$^T$ $\in \mathbb{R}^m$, satisfying $\sum_{i = 1}^{m} \lambda_i = 1$, $\lambda_i \geq 0$, \textit{i} = 1, ..., \textit{m}. Then the WS-transformation, giving rise to \boldsymbol{$\lambda'$}, on \boldsymbol{$\lambda$} can be defined as:

\begin{equation}
\begin{split}
\label{eq:ws-trans}
\boldsymbol{\lambda'} = WS(\boldsymbol{\lambda}) = (\frac{\frac{1}{\lambda_1}}{\sum_{i=1}^{m}{\frac{1}{\lambda_i}}}, \cdot\cdot\cdot, \frac{\frac{1}{\lambda_m}}{\sum_{i=1}^{m}{\frac{1}{\lambda_i}}})
\end{split}
\end{equation}
	
\section{Proposed Method}
\label{sec:prop-method}
	
    This section presents in details the proposed method for adapting weights during the evolutionary process called MOEA/D-URAW.

\subsection{Weights Generation}

MOEA/D-URAW uses the Uniformly Randomly method presented in \sref{sec:rand} for generating weight vectors. In this approach, the population size is flexible, that is, it independs on the number of objectives. However, maintain a fixed set of weights throughout the evolutionary process means assuming that the PF follows a given geometry. This is a problem because the use of a fixed set of weights does not guarantee to find well-distributed solutions for all PF shapes. Adapting weights helps to deal with this problem \cite{qi2014moea,li2017weights,wu2017adaptive}. An adaptation method is presented in \sref{sec:weights-adaptation}.

\subsection{Weights Adaptation}
\label{sec:weights-adaptation}
The method proposed in this paper is based on the methodology presented in the MOEA/D with Adaptive Weight Adjustment (MOEA/D-AWA)\cite{qi2014moea} that uses the sparsity level among individuals of the population to indicate which subproblem should be removed and which is apt to be added.

The sparsity level is based on the vicinity distance \cite{kukkonen2006fast}. This approach is defined in Equation \eqref{eq:sparsity}, where $L_{2}^{NN_{i}^j}$ is the $j$-th individual euclidean distance, $ind^j$, along with its $i$-th nearest neighbor of the population, $pop$. The $m$ closest euclidean distances are used, where $m$ is the number of objectives \cite{qi2014moea}.

\begin{equation}
\begin{split}
\label{eq:sparsity}
SL(ind^j, pop) = \prod_{i=1}^{m}{L_{2}^{NN_{i}^j}}
\end{split}
\end{equation}

The individual with the lowest level of sparsity or overcrowded is removed. That is, the sparsity level of each of the individuals in the population is calculated among the population itself using Equation \eqref{eq:sparsity}. If 5\% of the sub-problems (\textit{nus}) have not been removed yet of the population, then, repeat the process of calculating the sparsity levels and remove the most overcrowded subproblem \cite{qi2014moea}. An important observation is that the MOEA/D-URAW does not update the current evolutionary population by checking which individual is best suited to which weight vector as the MOEA/D-AWA does.

In addition to the evolutionary population, the MOEA/D framework has an external population (EP) that is used to store non-dominated solutions found during the search. The procedure to create new subproblems consists of calculating the sparsity level of each individual in an EP with respect to the population itself, using Equation \eqref{eq:sparsity}. Then, it generates a new subproblem using the individual $\boldsymbol{ind^{sp}}$ = (\boldsymbol{$x^{sp}$} \boldsymbol{$FV^{sp}$}) which has the highest sparsity level. The weight vector \boldsymbol{$\lambda^{sp}$} of the new constructed subproblem can be calculated as follows, in which \boldsymbol{$FV^{sp}$} = $(f_{1}^{sp}$ ... $f_{m}^{sp}$),

\begin{equation}
\begin{split}
\label{eq:add-subproblem}
\boldsymbol{\lambda^{sp}} = (\frac{\frac{1}{f_1^{sp} - z_1^*}}{\sum_{k=1}^{m}{\frac{1}{f_k^{sp} - z_k^*}}}, \cdot\cdot\cdot, \frac{\frac{1}{f_m^{sp} - z_m^*}}{\sum_{k=1}^{m}{\frac{1}{f_k^{sp} - z_k^*}}}), \prod_{j=1}^{m}{(f_j^{sp} - z_j^*)} \neq 0.
\end{split}
\end{equation}

At last, the solution of the new constructed subproblem as $ind^{sp}$ is set and added to the current population. The process is repeated until \textit{nus} subproblems are added to the population.

There is no consensus in the literature about the best moment for adaptation \cite{qi2014moea,li2017weights,wu2017adaptive}. In a preliminary evaluation the MOEA/D-URAW performed better when the weight update operation is conducted every 5\% of the total generations/evaluations. We followed it, and MOEA/D-URAW does not change the weight vectors during the last 10\% generations/evaluations.

\subsection{Algorithm Framework}
\label{sec:alg-framework}

\aref{alg:moeaduraw} presents the main procedure of MOEA/D-URAW. As can be seen, it has the same framework as MOEA/D version in \cite{li2009multiobjective}. However, it employs the weight vector update using sparsity level (line 26-37), initialization of weights using uniformly randomly method (line 2) and the limitation of the size of EP (line 24-25).

\begin{algorithm}
\caption{MOEA/D-URAW}
\label{alg:moeaduraw}
EP $\gets \emptyset$;

Initialize the population $P$ and a set of weight vectors $W$;

Apply the WS-transformation on the weight vectors $W$;

Determine the neighbors of each weight vector of $W$;

Calculate the reference point according to $P$;

Gen $\gets$ 0;

\While{ Gen $<$ $Gen_{max}$}
{
	\For{$each$ $i \in \{1,...,N\}$}
    {
    	\If{$uniform(0, 1) < \delta$}
        { 
        	$E \gets B(i)$;
        }
        \Else
        {
        	$E \gets \{1,...,N\}$;
        }
        Randomly select mating solutions from $E$ to generate an offspring $\bar{\textbf{x}}$, Evaluate \textbf{F}($\bar{\textbf{x}}$);
        
        $update \gets 0$;
        
        \While{ $update < nr$ and $E \neq \emptyset$}
        {
        	$j \gets$ Randomly select an index from $E$;
            
            $E \gets E$  $\backslash$  $j$;
            
            \If{$g^{TCH}(\bar{\textbf{x}} \vert \boldsymbol{w}^j , \boldsymbol{z^*})$ $\leq$ $g^{TCH}(\boldsymbol{x}^j \vert \boldsymbol{w}^j , \boldsymbol{z^*})$ }
            {
            	$\textbf{x}^j$ $\gets$ $\bar{\textbf{x}}$;
                
                $update \gets update + 1 $;
            }
        }
        
		\If{$\nexists$q $\in$ EP, q $\prec$ $\bar{\textbf{x}}$}
        { 
        	$EP \gets EP$ $\cup$ $\bar{\textbf{x}}$;
            
            $EP \gets EP$ $\backslash$ $\{ q \in EP$  $\vert$ $\bar{\textbf{x}}$ $\prec q \}$;
        }
	}
    
    \If{$\vert$EP$\vert$ > 2$\vert$P$\vert$}
    { 
		Remove from $EP$ the individual with the highest sparsity level;
	}  
    
	\If{Gen = $Gen_{max}$ $\times$ 5\% \textbf{and} $< Gen_{max}$ $\times$ 90\%}
    { 
    
    	$adjust \gets 0$
    
    	\While{$adjust$ $<$ $nus$}
        {        	
            Calculate the sparsity level of each individual in population $P$ among $P$ by Equation \eqref{eq:sparsity};
            
         	Remove the individual with the minimum sparsity level;   
            
        	$adjust$ $\gets$ $adjust$ $+$ $1$;
        }
    
		\While{$adjust$ $>$ $0$}
        {
        	Calculate the sparsity level of each individual in $EP$ among the population $P$ by Equation \eqref{eq:sparsity};
            
            Generate a new subproblem using the individual which has the largest sparsity level by Equation \eqref{eq:add-subproblem};

			Add the newly constructed subproblem associated with the individual which has the largest sparsity level to the current population $P$;
                      
        	$adjust$ $\gets$ $adjust$ $-$ $1$;
        
        }
        
		Update the neighbors of each weight vector of \textit{W};
	}      
    
    $Gen \gets Gen + 1$;
}
\textbf{return} \textit{P};
\end{algorithm}

\section{Experimental Setup}
\label{sec:setup}

\subsection{Test Problems}

In the experimental study, we used eight modified WFG4 test problems, i.e., WFG41 to WFG48 \cite{wang2015preference}, from 2 to 6 objectives. They have different PF characteristics, namely continuous and discontinuous, convex, concave, strong convex, strong concave and mixed PFs with different shapes. Each WFG problem had 100 executions. The number of decision variables is defined as $n = k + l$, and $m$ determines the number of objectives. The others settings are described in \tref{tab:wfg41-48-params}. Note that, the population size was thus defined, in order to have a fair comparison between the test algorithms, since Das and Dennis method does not provide a flexible value when treating many-objectives.

    \begin{table}[!ht]
  \renewcommand{\arraystretch}{1.3}	
  \caption{Settings used for WFG41 to WFG48.}
  \label{tab:wfg41-48-params}
  \centering
  \begin{tabular}{cc}
    \toprule
   		Parameters&Values\\
    \midrule
    	$runs$ & 100\\
    	$maxGen$ & 400\\
        $N$ (2-4 objectives) & 120\\
        $N$ (5 and 6 objectives) & 126\\
        $T$ & 24\\        
        $nus$ (0.05$N$) & 6\\
        $k$ (2 objectives) & 2\\
        $k$ (otherwise) & $m$-1\\
        $l$ & 10\\
  \bottomrule
\end{tabular}
\end{table}

\subsection{Test Algorithms}
\label{sec:test-alg}

To assess the performance of MOEA/D-URAW, the Wilcoxon's rank sum test is used with the significant level of 5\%. Three ways of initializing unadjusted weights in the MOEA/D framework presented in section \sref{sec:alg-framework} is considered for comparison: Das and Dennis \cite{das1998normal}, Uniformly Randomly \cite{zhang2009performance} and TCH Scalarizing Function \cite{gu2012multiobjective,li2017weights}. These algorithms used the same general settings as shown in \tref{tab:general-params}.

	\begin{table}[!ht]
  \renewcommand{\arraystretch}{1.3}	
  \caption{General Parameters.}
  \label{tab:general-params}
  \centering
  \begin{tabular}{cc}
    \toprule
   		Parameters&Values\\
    \midrule
        Crossover &  Differential Evolution\\ 
        $CR$ & 1.0\\
        $F$ & 0.5\\
        Mutation &  Polynomial\\
        $P_m$ & 1/$n$\\
        $\eta_m$ & 20\\
        $\delta$ & 0.9\\
        $nr$ & 2\\
  \bottomrule
\end{tabular}
\end{table}

\subsection{Performance Metric}

For WFG41 to WFG48 test problems, we use the Hypervolume metric (HV), since the PFs are unknown. Given a reference point $z^r$ = ($z_1^r$, ... , $z_n^r$)$^T$ dominated by all Pareto-optimal solutions, the HV of a solution set \textit{P} is defined as the volume of the objective space dominated by all solutions in \textit{P}, bounded by $z^r$:
    \begin{equation}
    \label{hypervolume}
    HV(P) = VOL(\bigcup_{z \in P} [z_1, z_1^r] \times ... \times [z_m, z_m^r]),
    \end{equation}
where VOL indicates the Lebesgue measure. The objective vectors of the final solution set are normalized according to min-max normalization concerning all experiments before calculating the HV with $z^r$ = (1.2, ... , 1.2)$^T$.

\section{Experimental Results}
\label{sec:studies}

	As shown in \sref{sec:setup}, the Wilcoxon's rank sum test is used to indicate the MOEA/D-URAW performance when compared to the DD, UR and TSF approaches in MOOP and MaOP. According to \tref{tab:mopResults} and \tref{tab:maopResults}, MOEA/D-URAW presents results with a significance level of 5\%, better in 72.5\% of cases, draws at 5\% and not so good at 22.5\%. Therefore, the proposed method was better or equal to in 77.5\% of the problems. The results in the context of MOOP and MaOP are detailed below.

Firstly, \tref{tab:mopResults} and \tref{tab:mopMedianBestResults} shows the HV results for 2 and 3 objectives. In this context of MOOP, the MOEA/D-URAW  performs the best on most of the test instances except for WFG47 for 2 objectives, and WFG43 for 3 objectives. For these cases, Tchebycheff Scalarizing Function, and Das and Dennis obtained the best HV results, respectively. 

\begin{figure*}[ht!]
  \centering
  \subfigure[WFG41 - DD]{\includegraphics[width=0.245\linewidth]{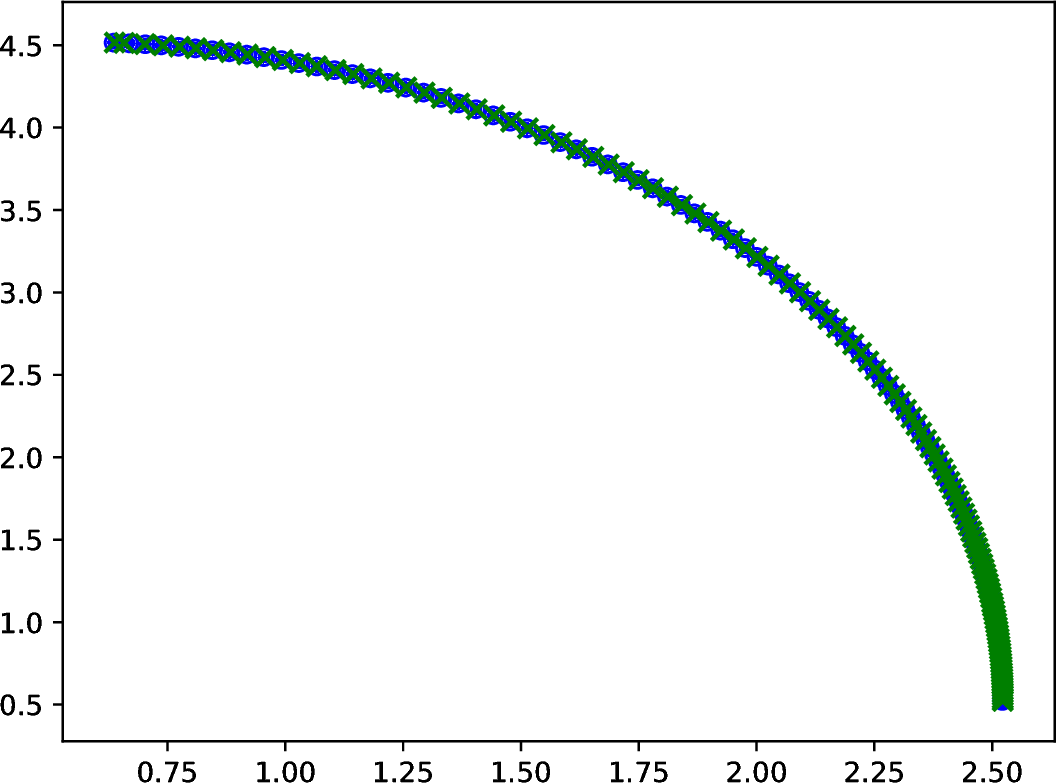}}
  \subfigure[WFG41 - UR]{\includegraphics[width=0.245\linewidth]{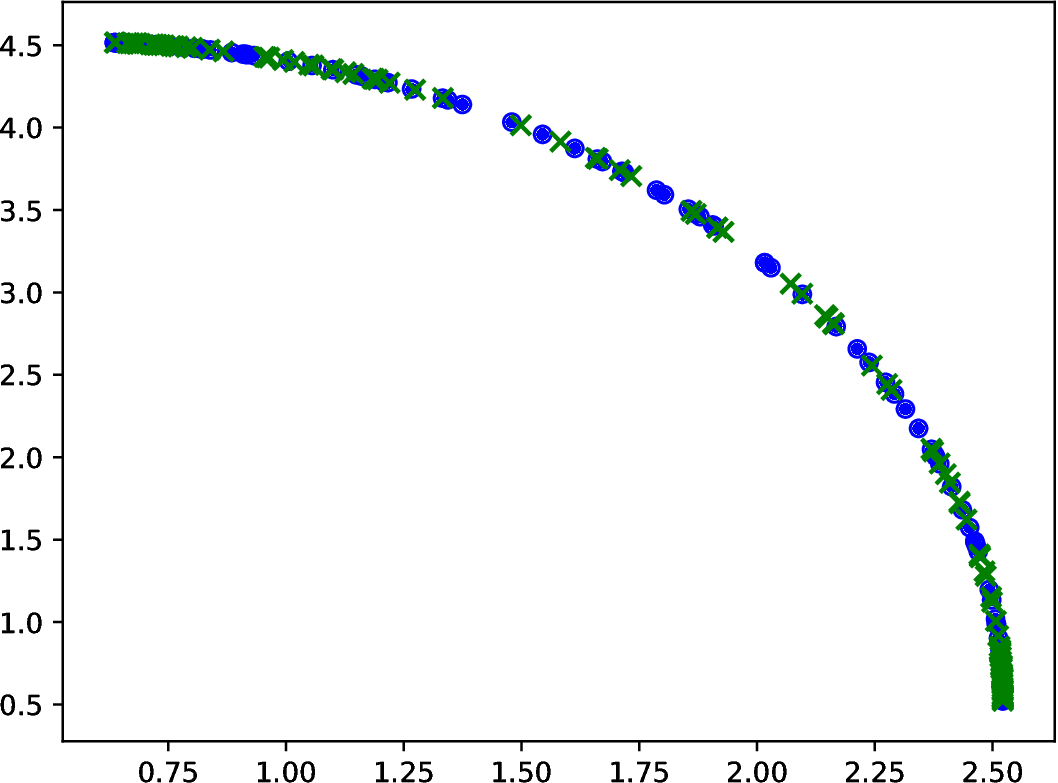}}
  \subfigure[WFG41 - TSF]{\includegraphics[width=0.245\linewidth]{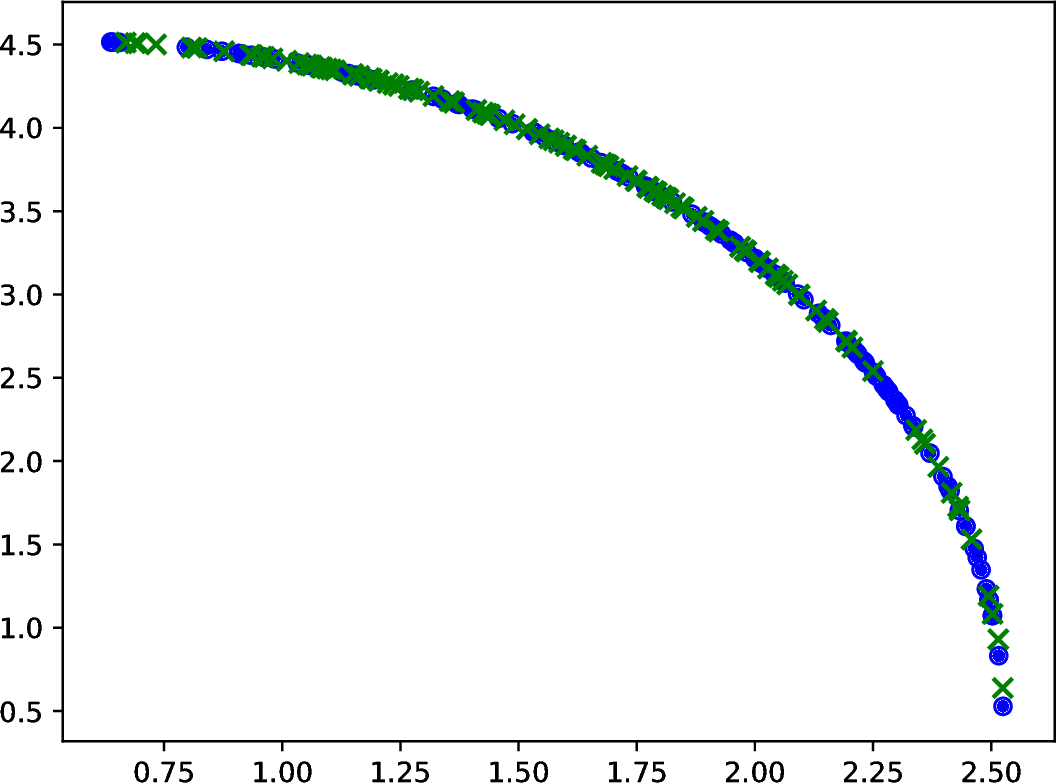}}
  \subfigure[WFG41 - URAW]{\includegraphics[width=0.245\linewidth]{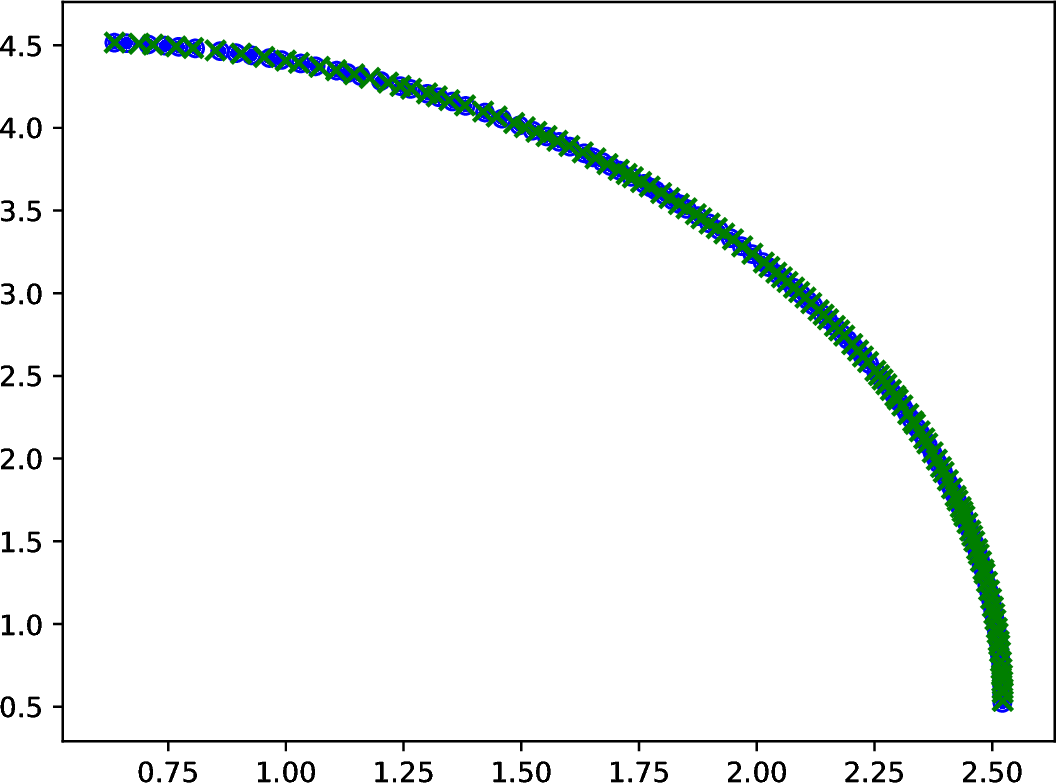}}
  
  \centering
  \subfigure[WFG42 - DD]{\includegraphics[width=0.245\linewidth]{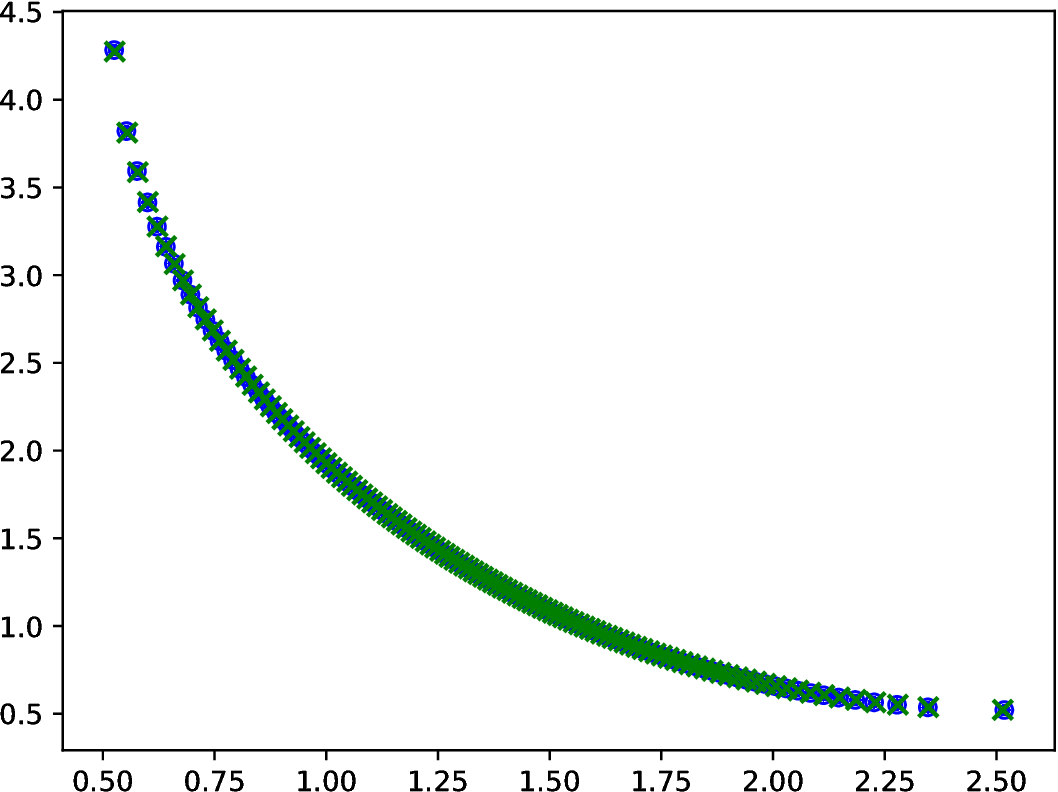}}
  \subfigure[WFG42 - UR]{\includegraphics[width=0.245\linewidth]{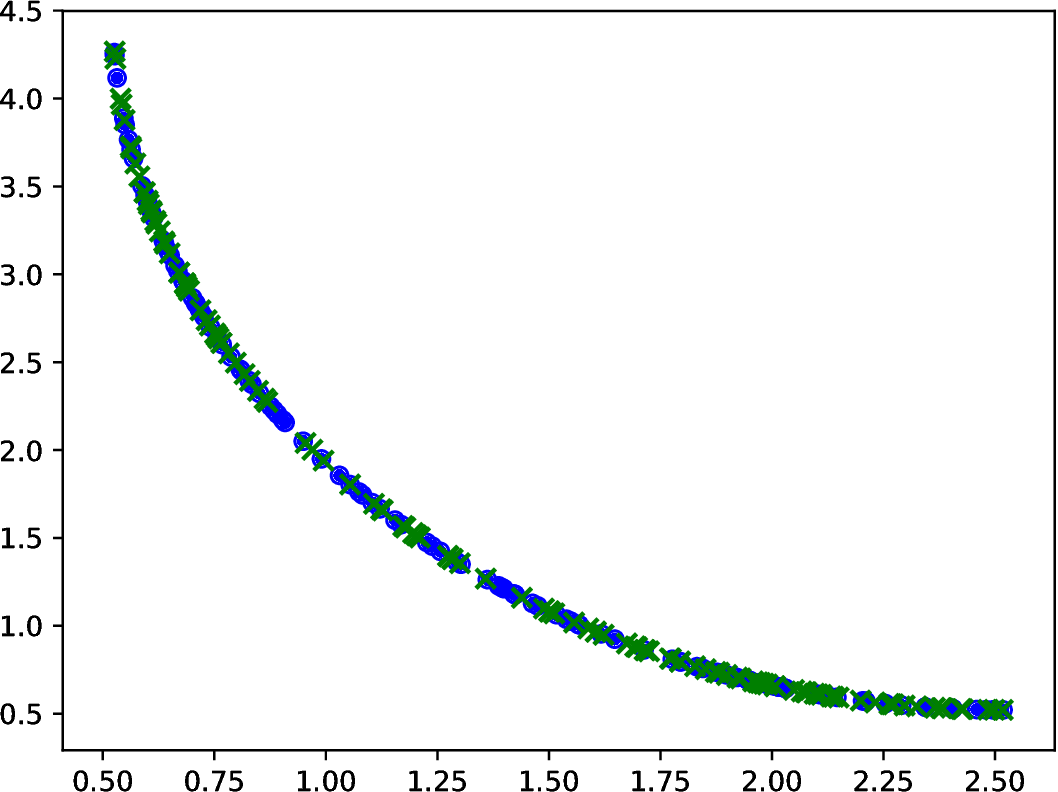}}
  \subfigure[WFG42 - TSF]{\includegraphics[width=0.245\linewidth]{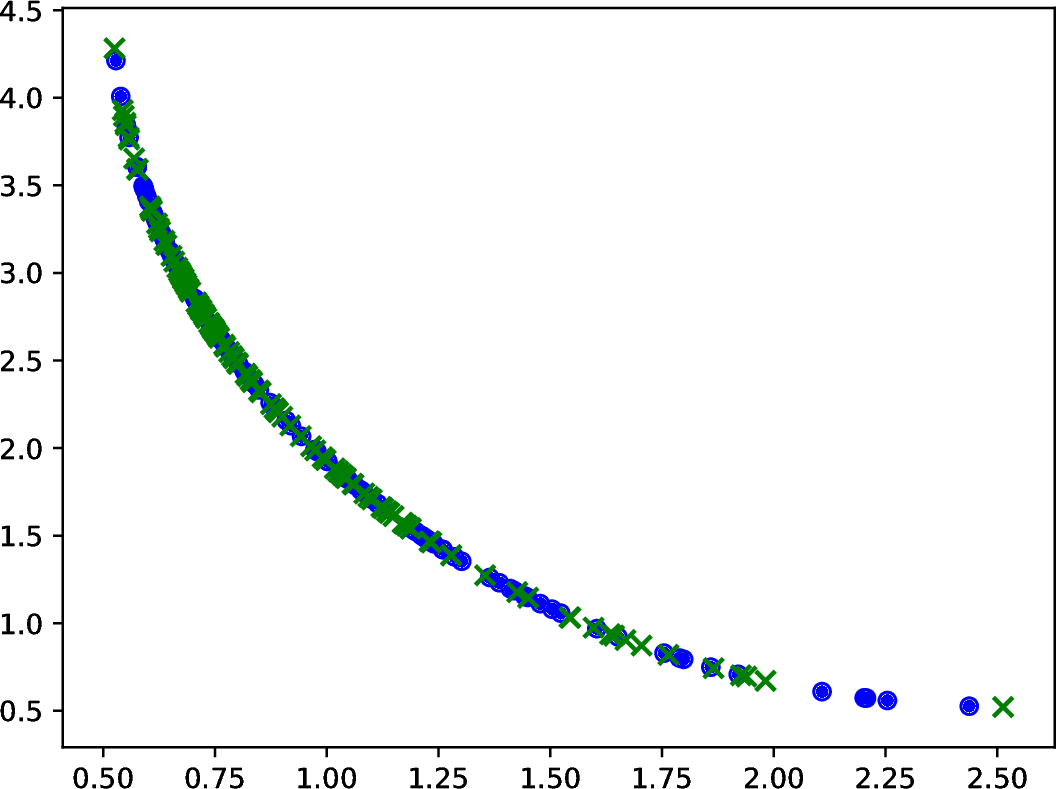}}
  \subfigure[WFG42 - URAW]{\includegraphics[width=0.245\linewidth]{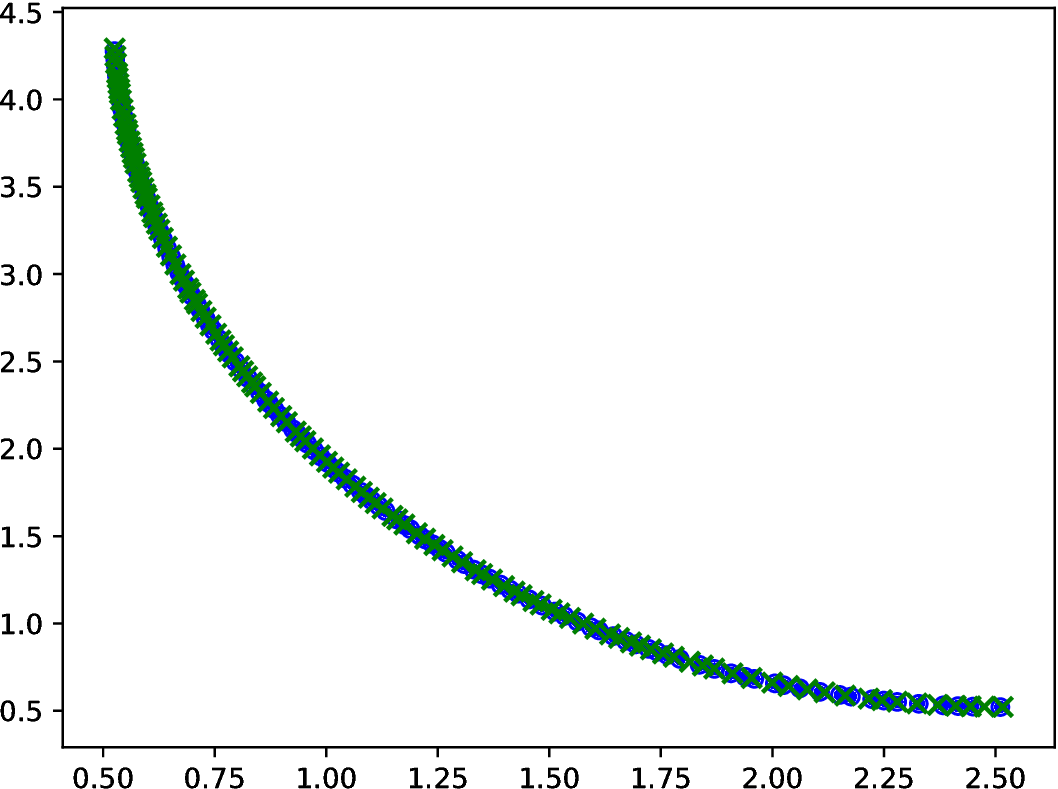}}
  
  \centering
  \subfigure[WFG43 - DD]{\includegraphics[width=0.245\linewidth]{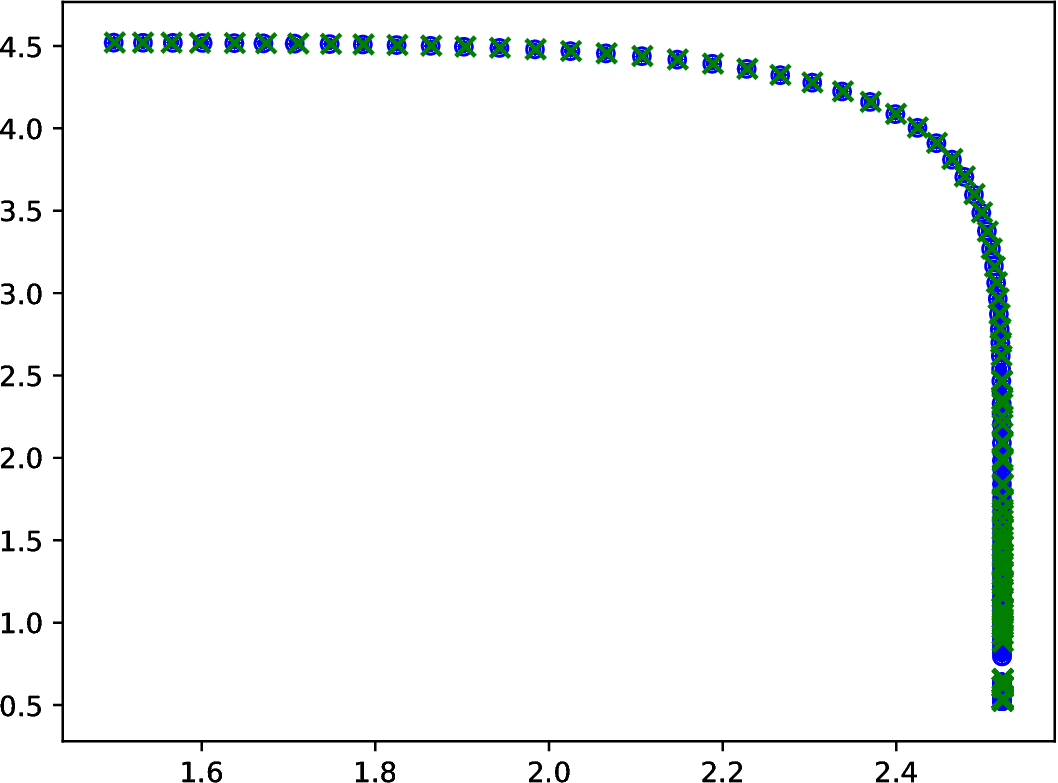}}
  \subfigure[WFG43 - UR]{\includegraphics[width=0.245\linewidth]{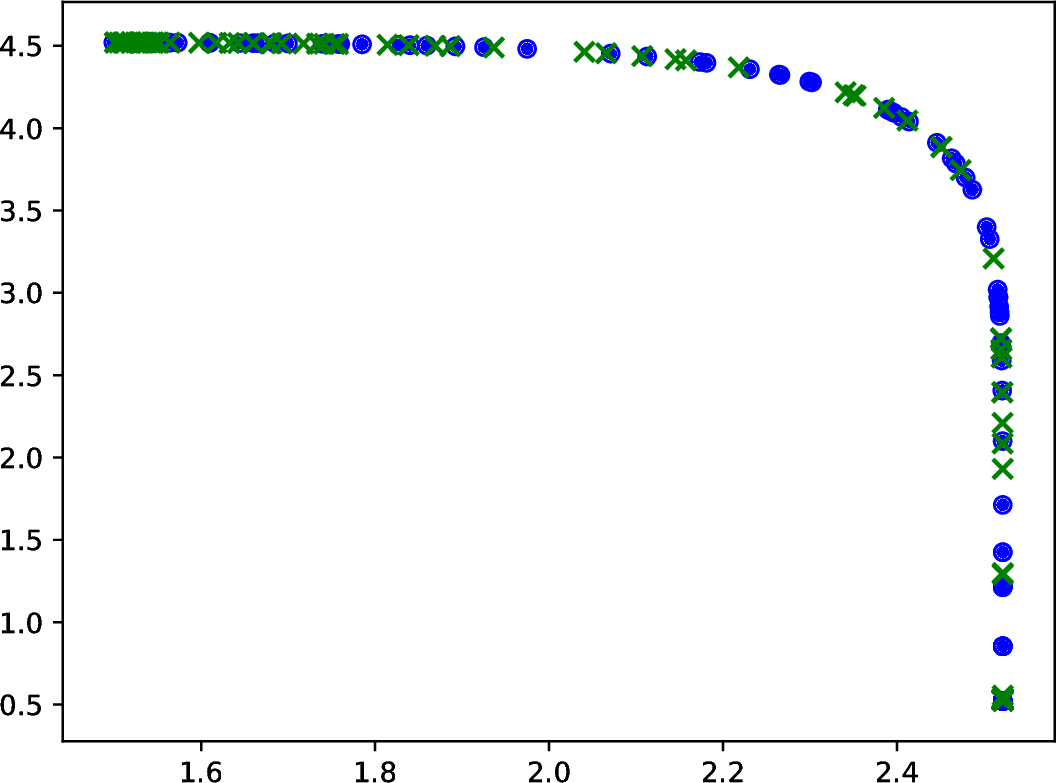}}
  \subfigure[WFG43 - TSF]{\includegraphics[width=0.245\linewidth]{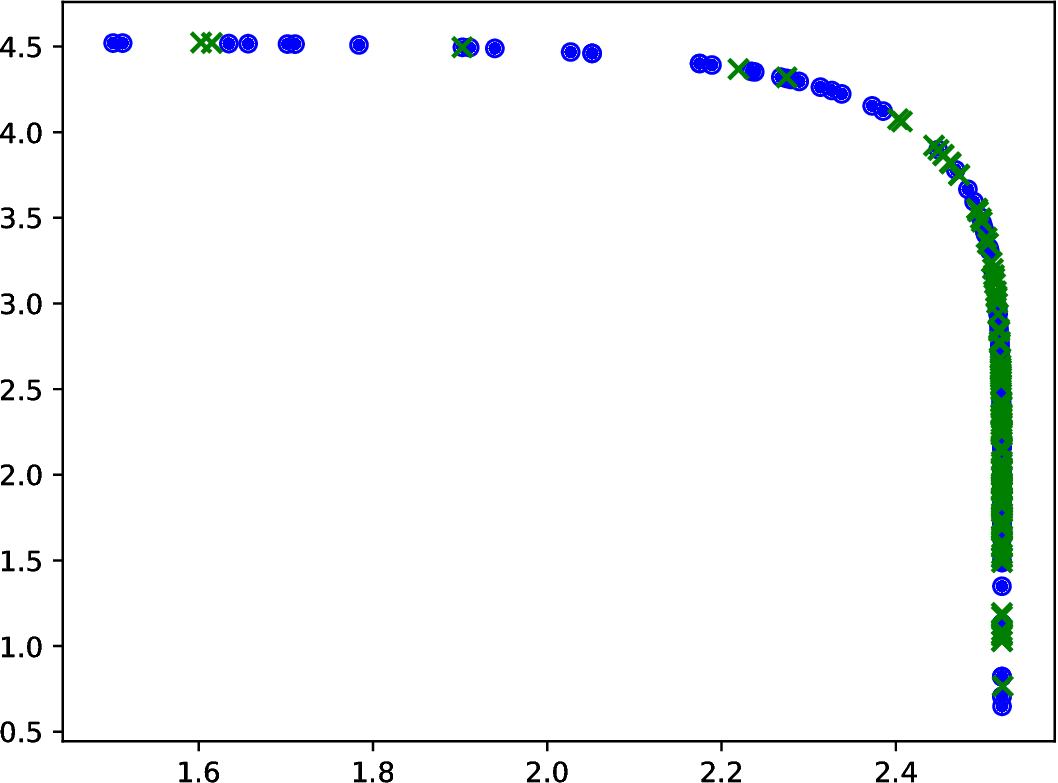}}
  \subfigure[WFG43 - URAW]{\includegraphics[width=0.245\linewidth]{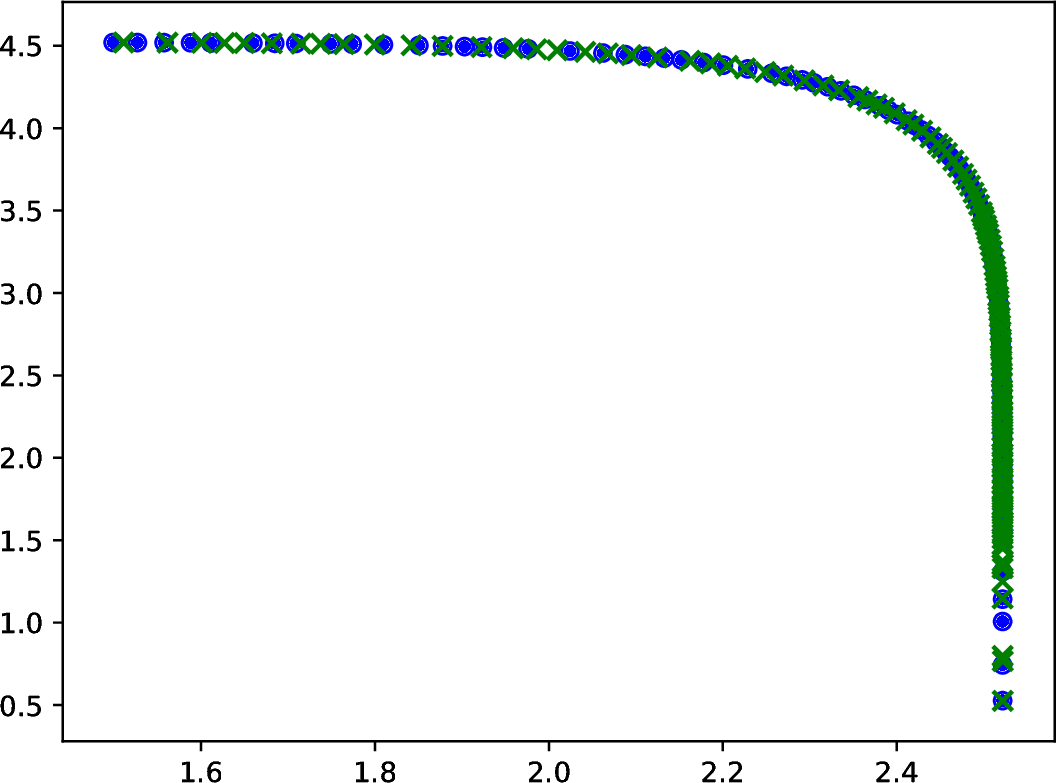}}
  
  \centering
  \subfigure[WFG44 - DD]{\includegraphics[width=0.245\linewidth]{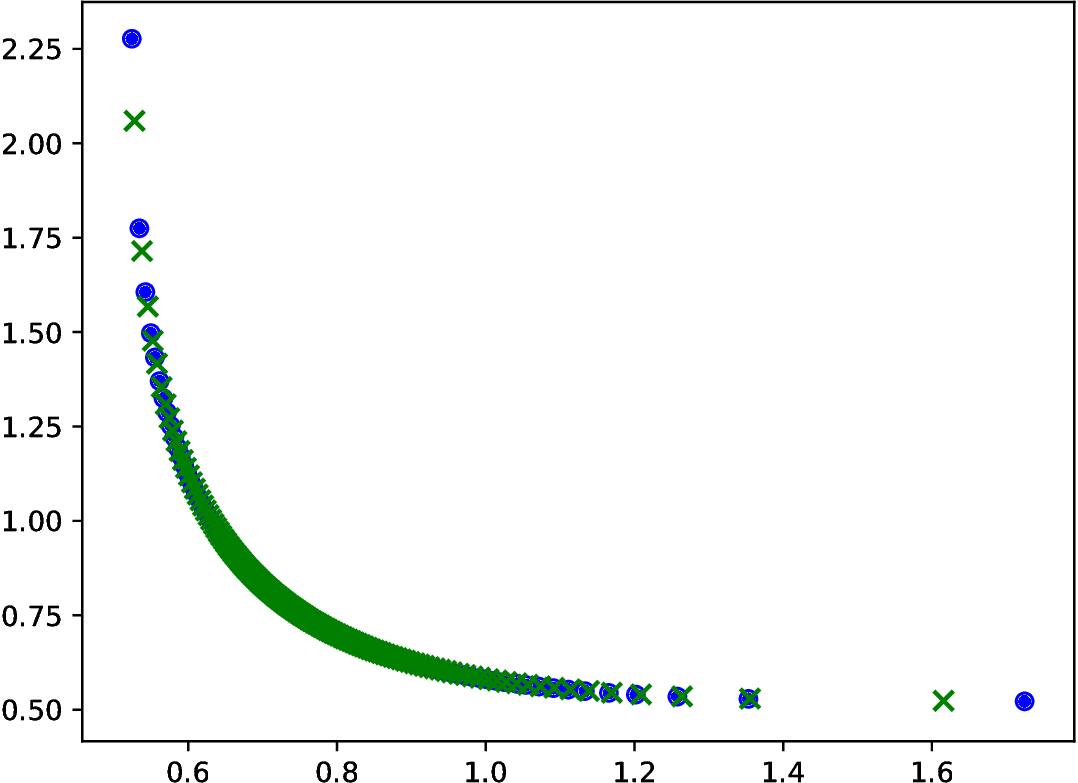}}
  \subfigure[WFG44 - UR]{\includegraphics[width=0.245\linewidth]{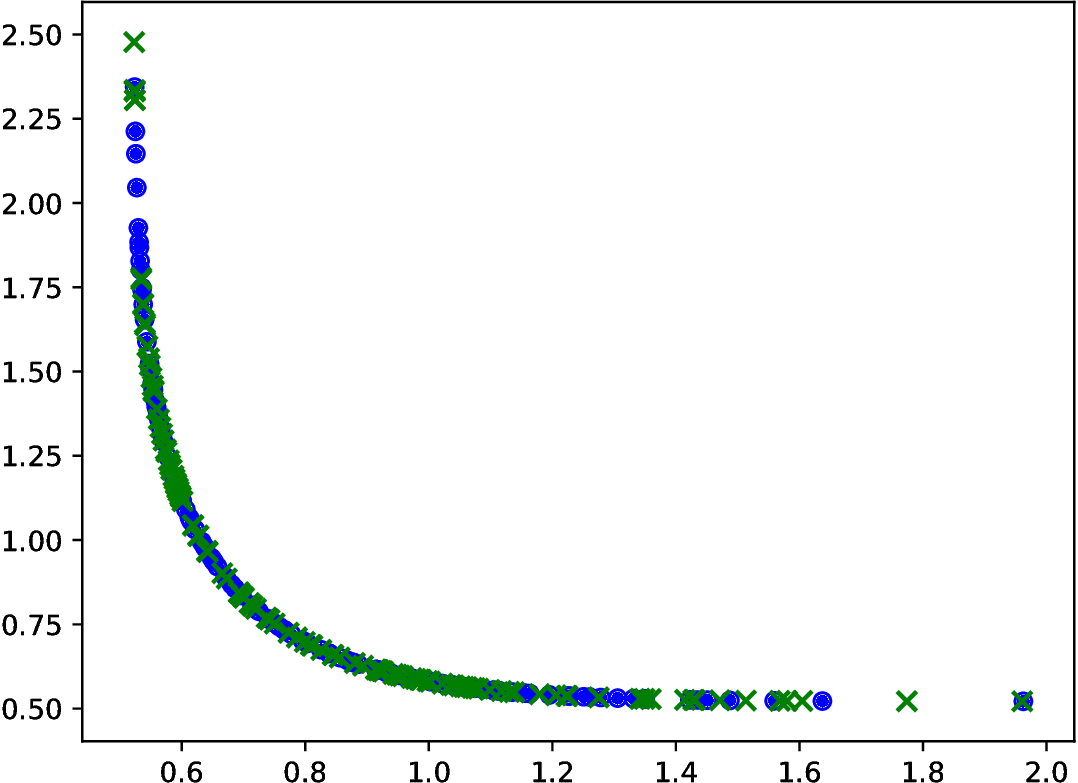}}
  \subfigure[WFG44 - TSF]{\includegraphics[width=0.245\linewidth]{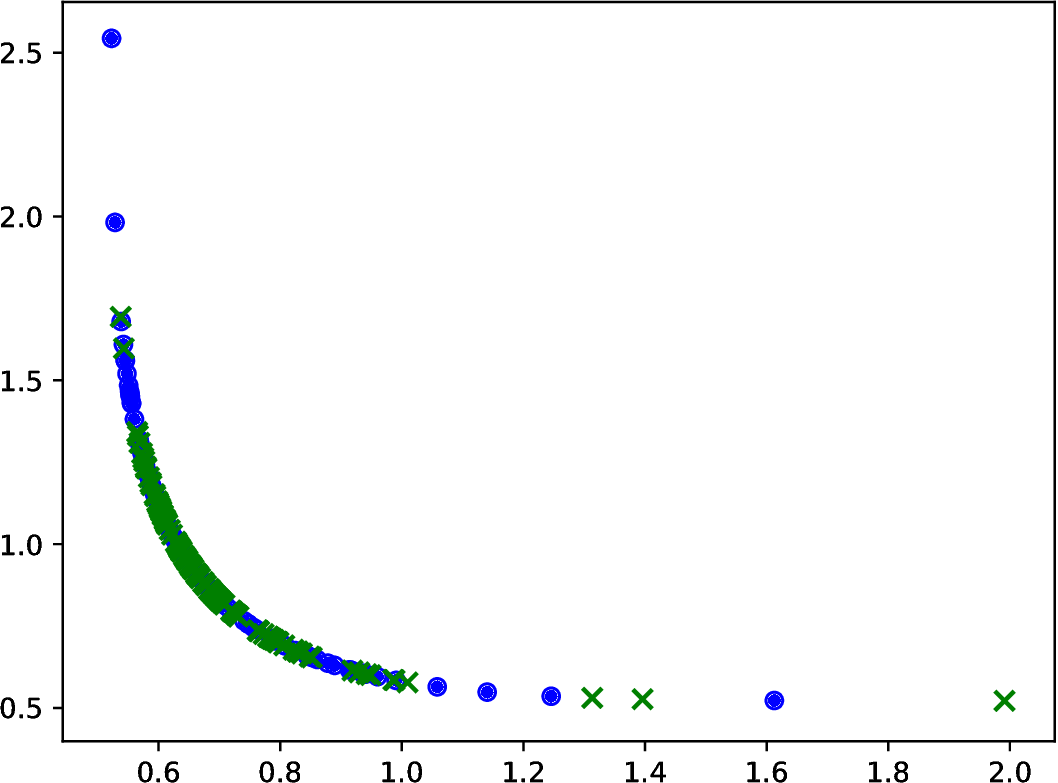}}
  \subfigure[WFG44 - URAW]{\includegraphics[width=0.245\linewidth]{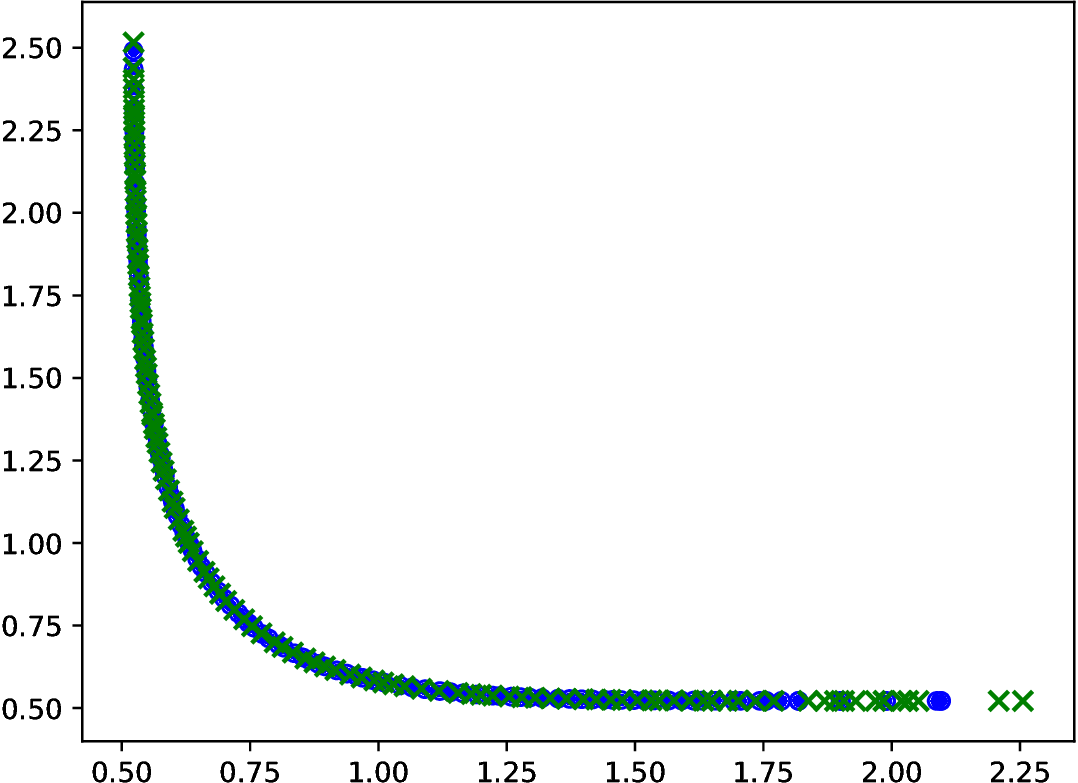}}
  
  \caption{Final solutions set with the best (blue circle) and median (green cross sign) HV metric values obtained by the algorithms on WFG41 to WFG44.}
  \label{fig:wfg41-44}
\end{figure*}

\begin{figure*}[ht!]
  \centering
  \subfigure[WFG45 - DD]{\includegraphics[width=0.245\linewidth]{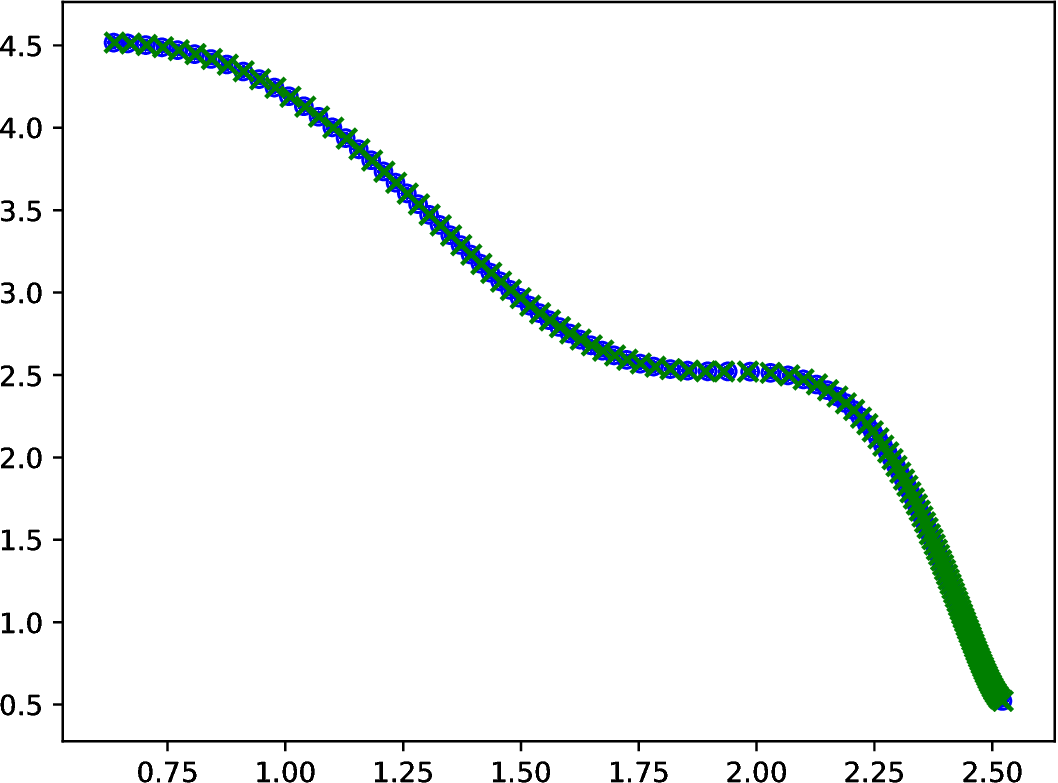}}
  \subfigure[WFG45 - UR]{\includegraphics[width=0.245\linewidth]{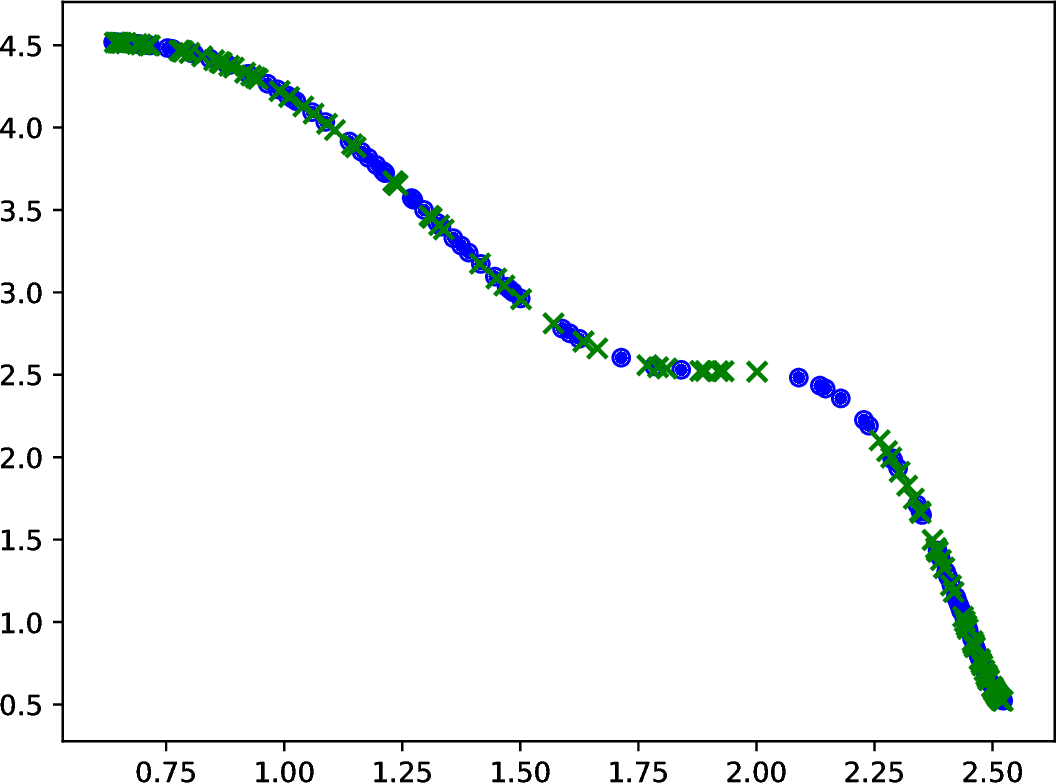}}
  \subfigure[WFG45 - TSF]{\includegraphics[width=0.245\linewidth]{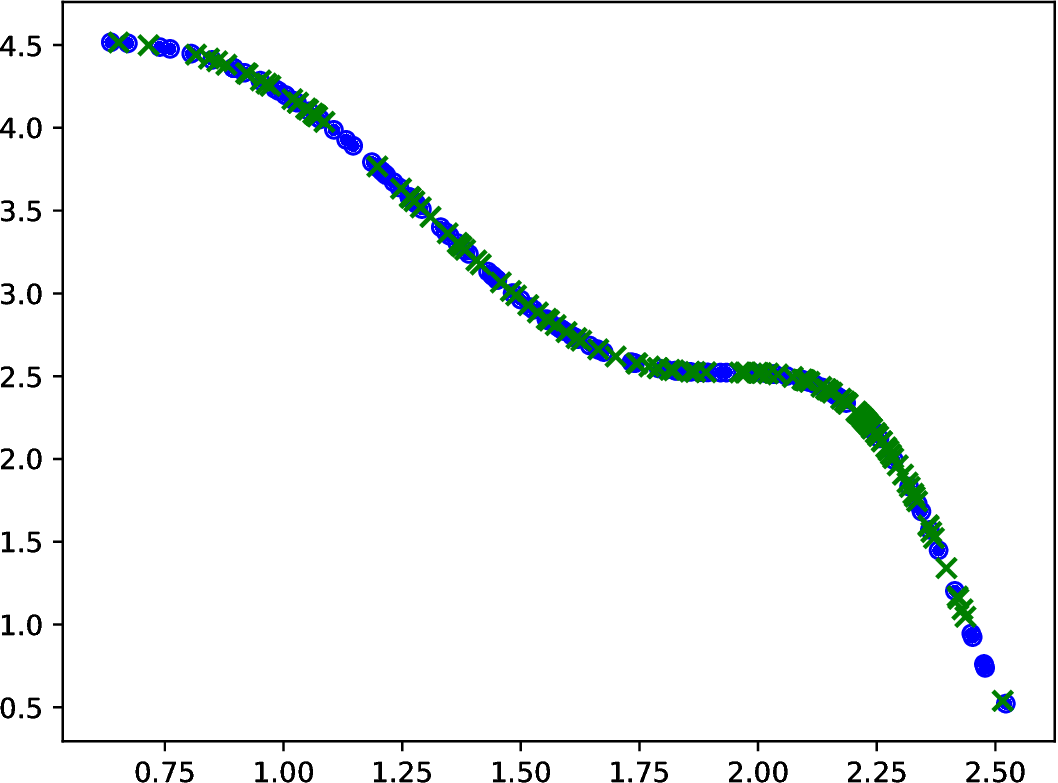}}
  \subfigure[WFG45 - URAW]{\includegraphics[width=0.245\linewidth]{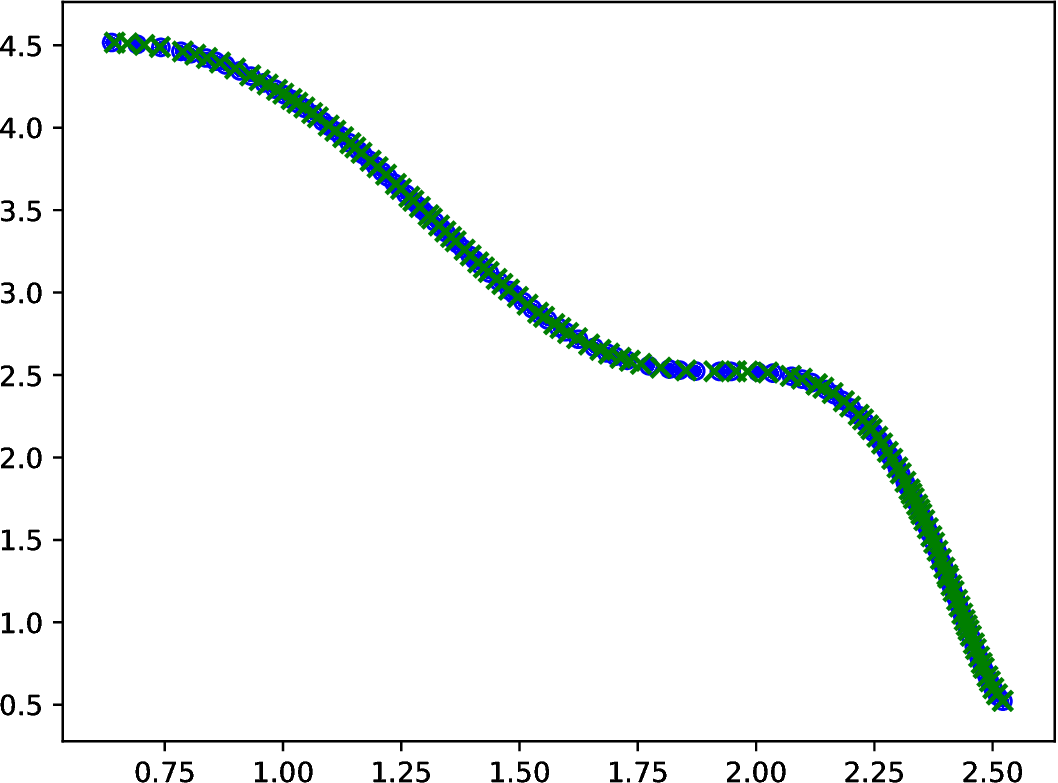}}
  
  \centering
  \subfigure[WFG46 - DD]{\includegraphics[width=0.245\linewidth]{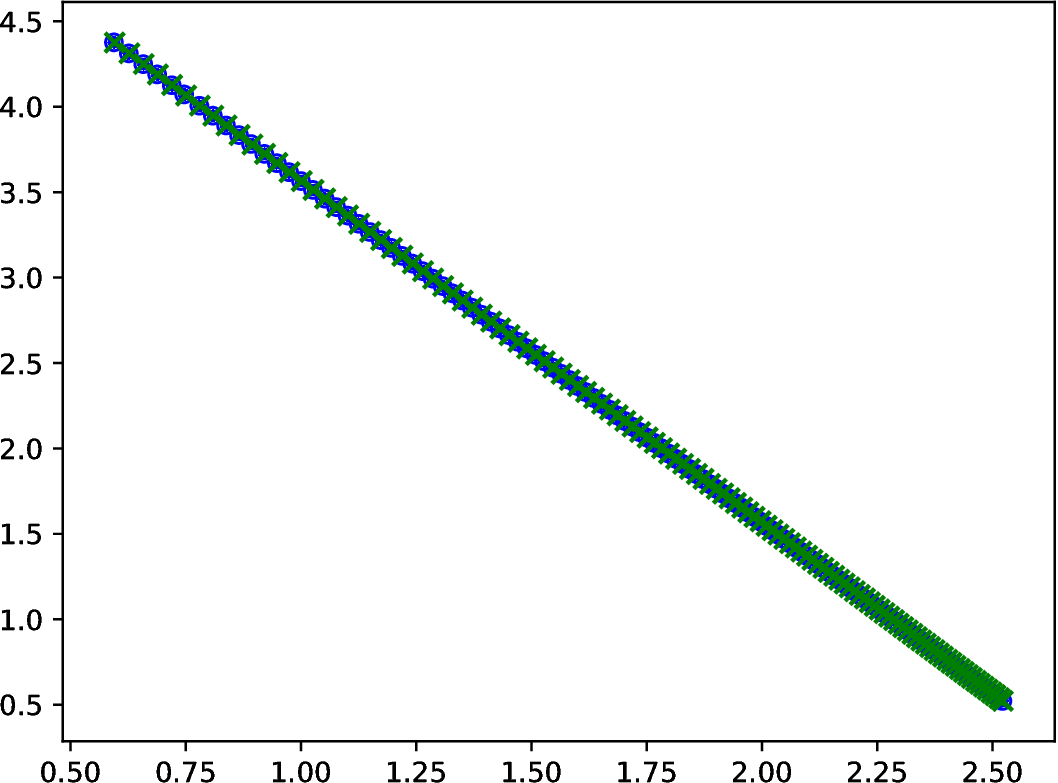}}
  \subfigure[WFG46 - UR]{\includegraphics[width=0.245\linewidth]{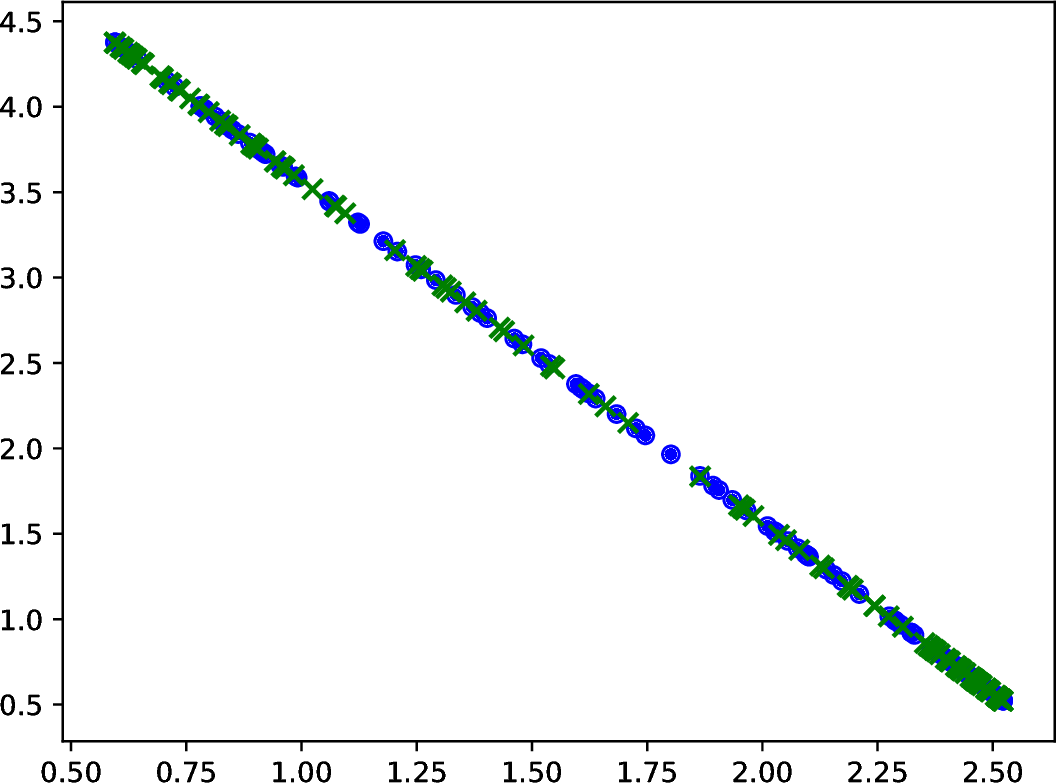}}
  \subfigure[WFG46 - TSF]{\includegraphics[width=0.245\linewidth]{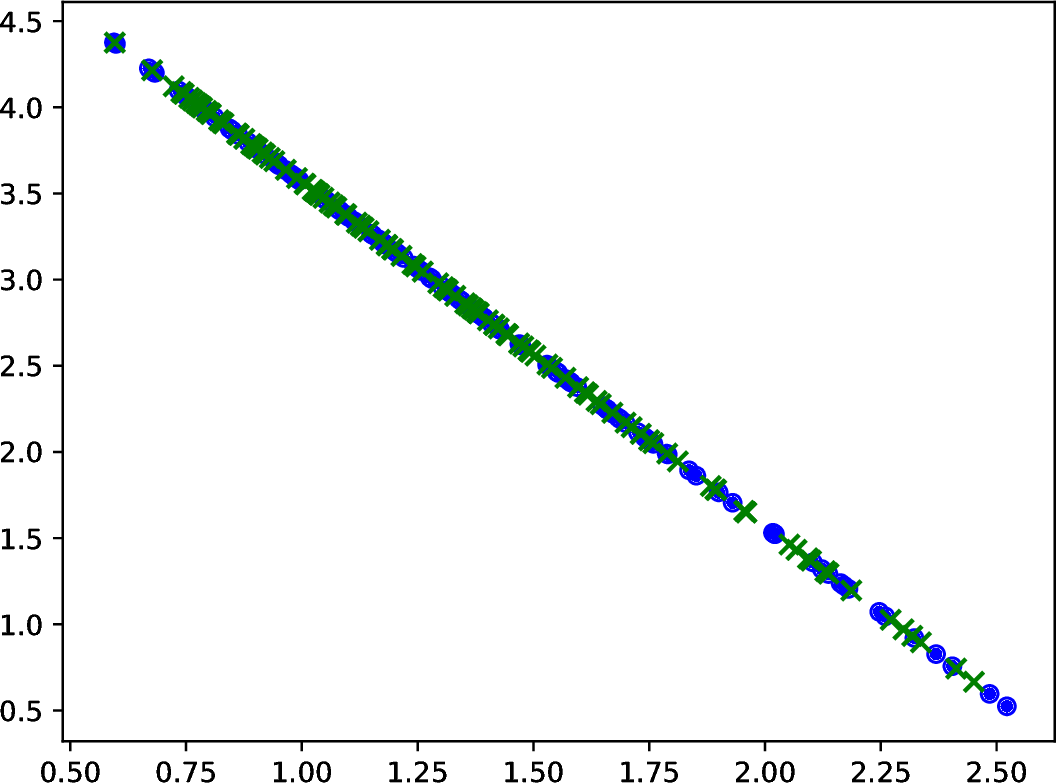}}
  \subfigure[WFG46 - URAW]{\includegraphics[width=0.245\linewidth]{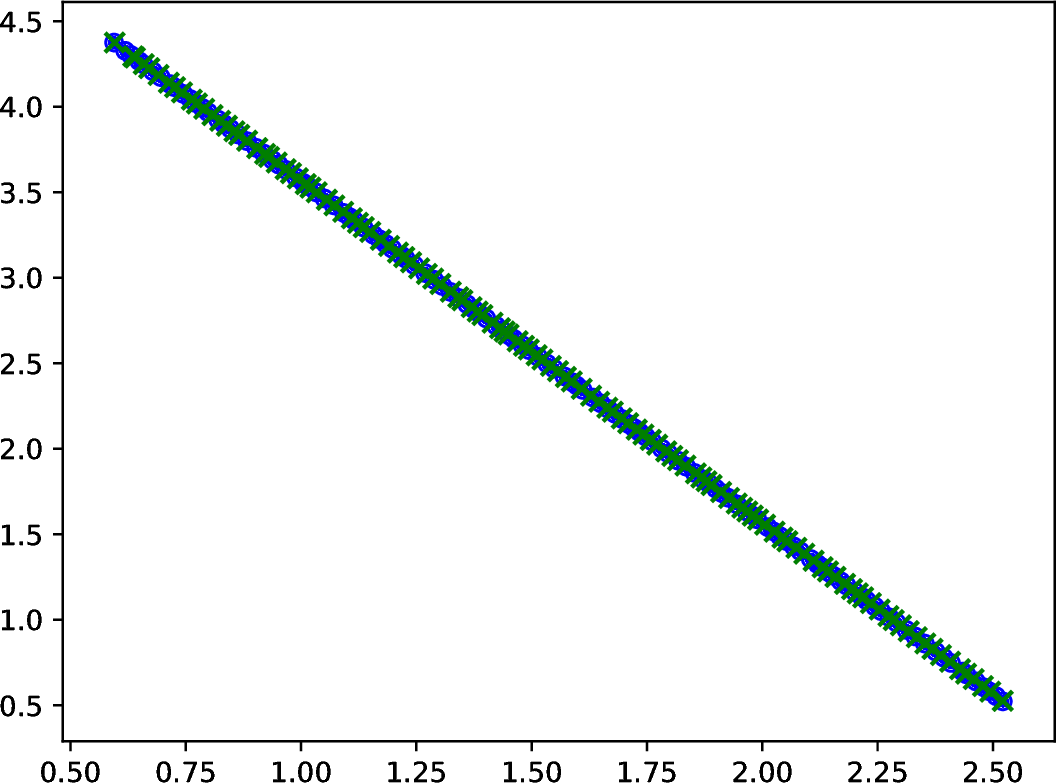}}
  
  \centering
  \subfigure[WFG47 - DD]{\includegraphics[width=0.245\linewidth]{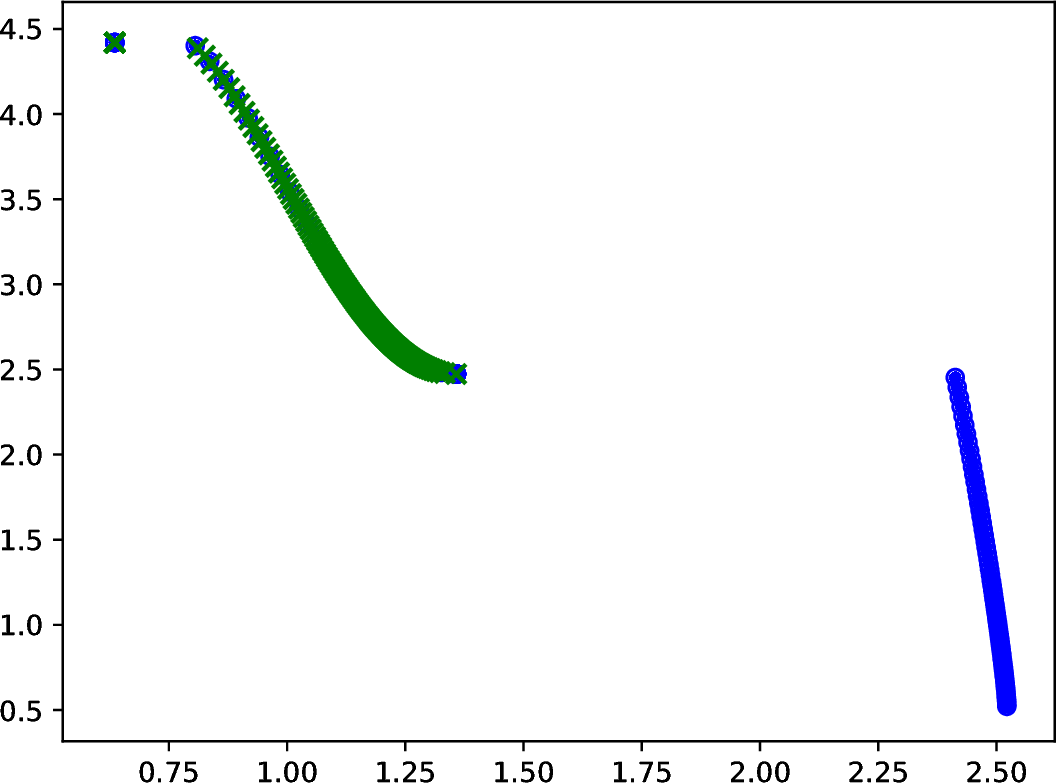}}
  \subfigure[WFG47 - UR]{\includegraphics[width=0.245\linewidth]{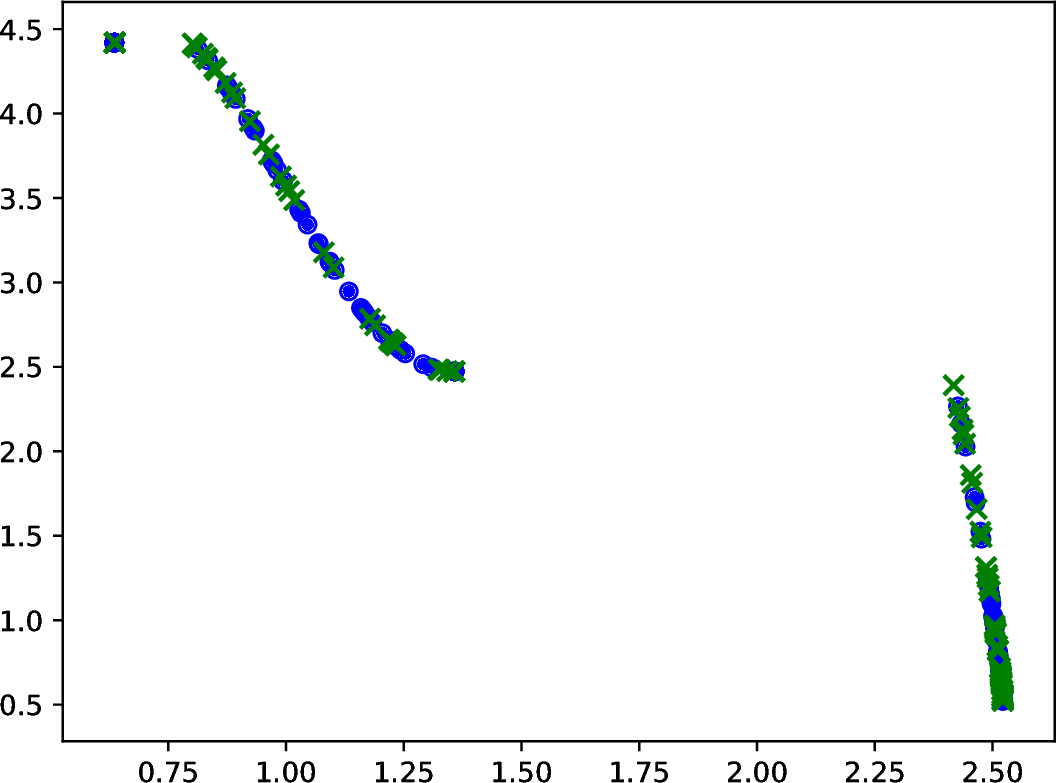}}
  \subfigure[WFG47 - TSF]{\includegraphics[width=0.245\linewidth]{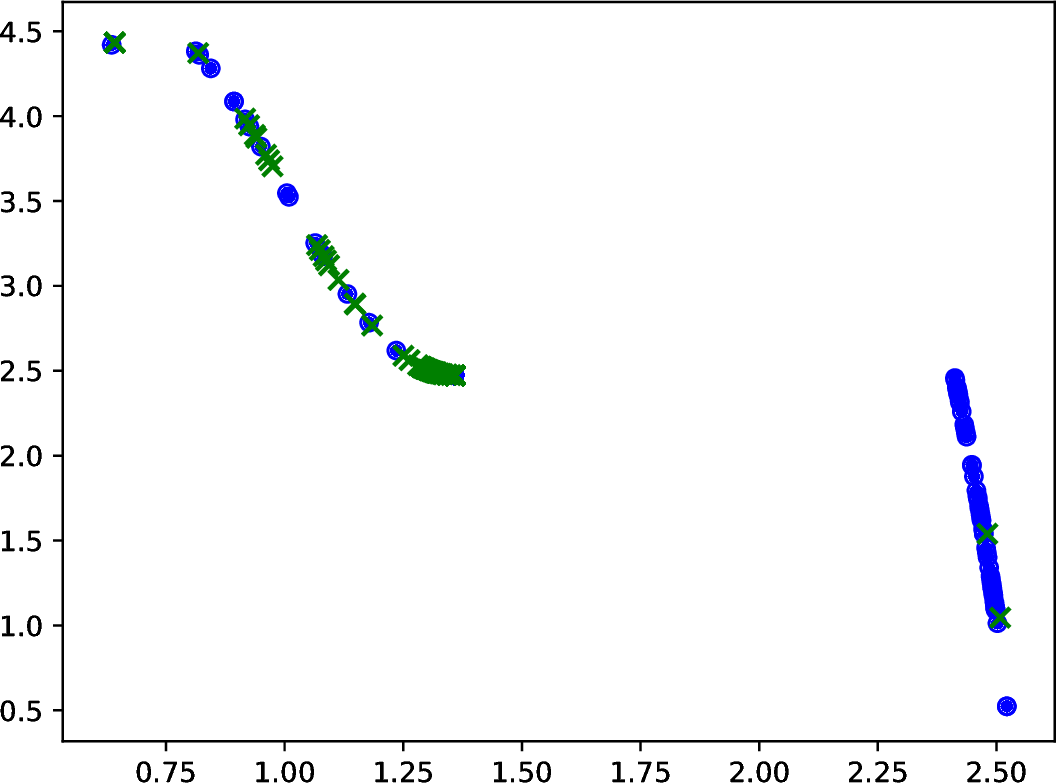}}
  \subfigure[WFG47 - URAW]{\includegraphics[width=0.245\linewidth]{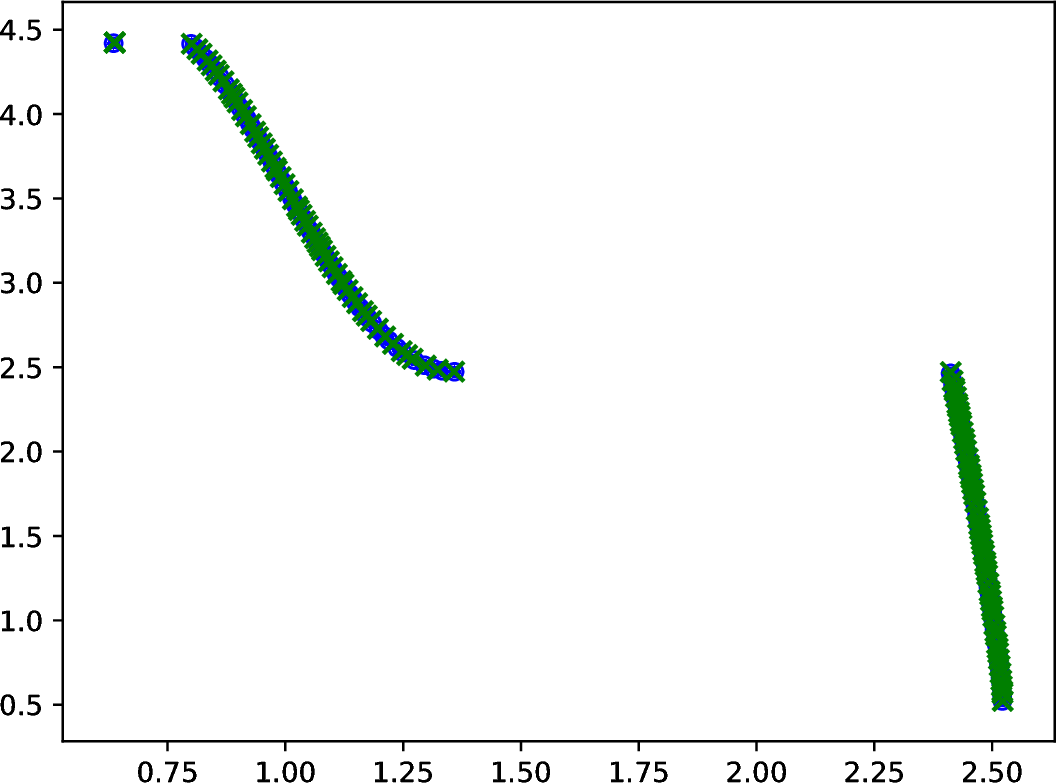}}
  
  \centering
  \subfigure[WFG48 - DD]{\includegraphics[width=0.245\linewidth]{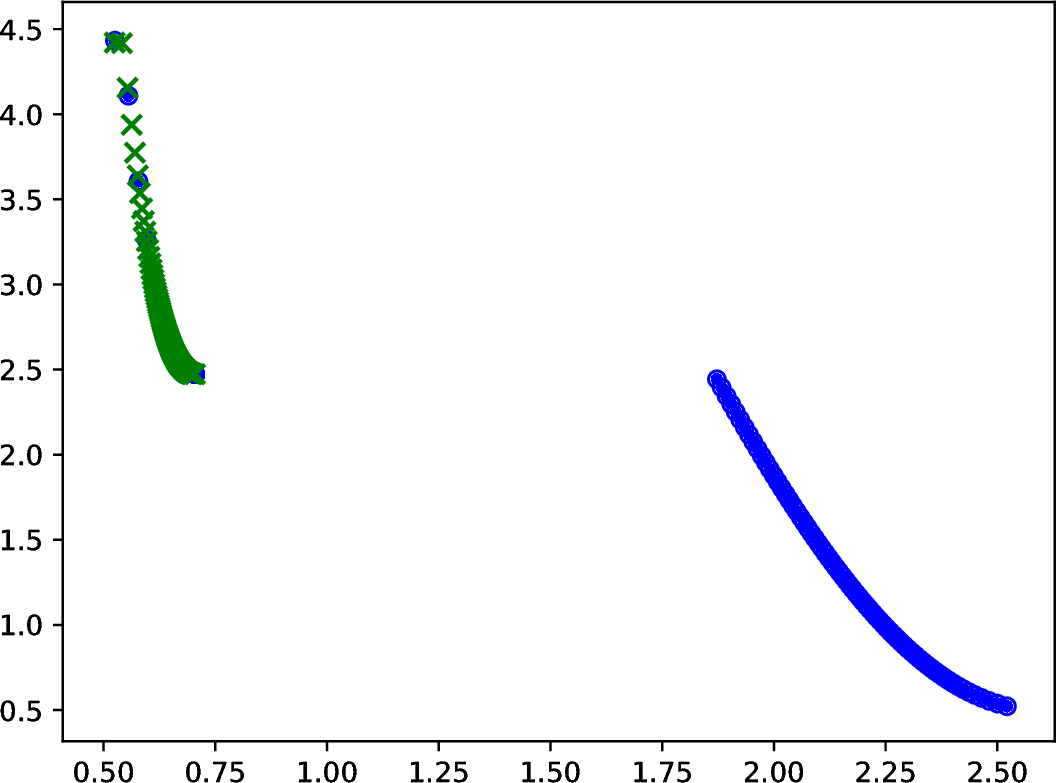}}
  \subfigure[WFG48 - UR]{\includegraphics[width=0.245\linewidth]{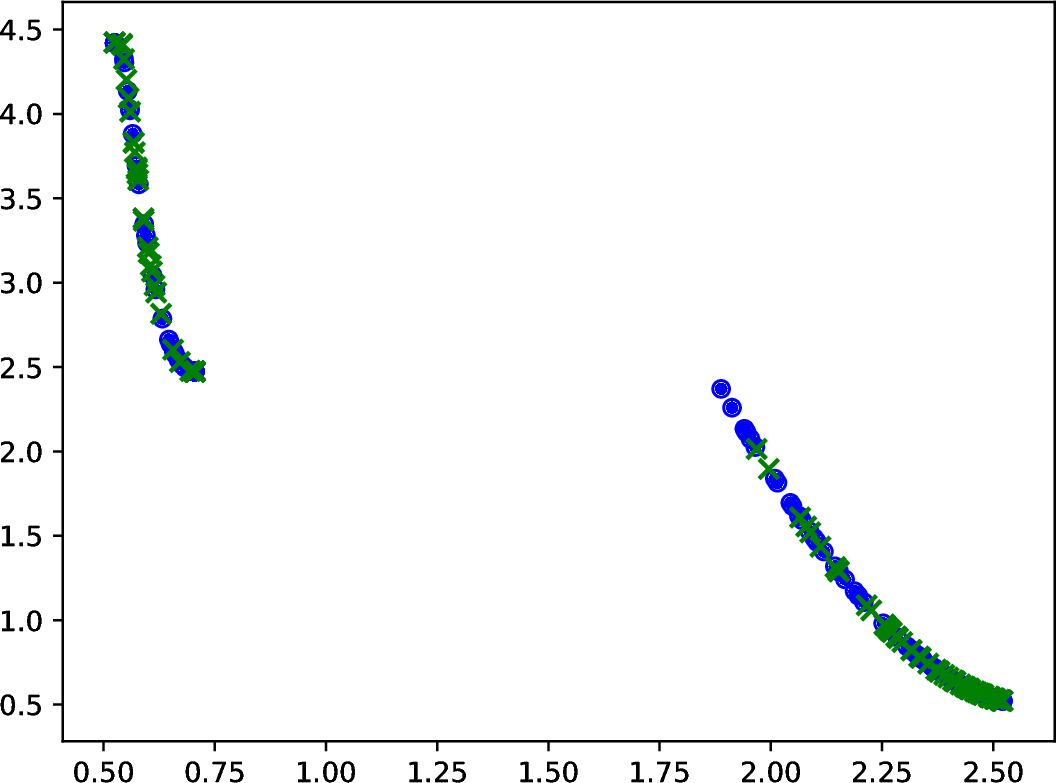}}
  \subfigure[WFG48 - TSF]{\includegraphics[width=0.245\linewidth]{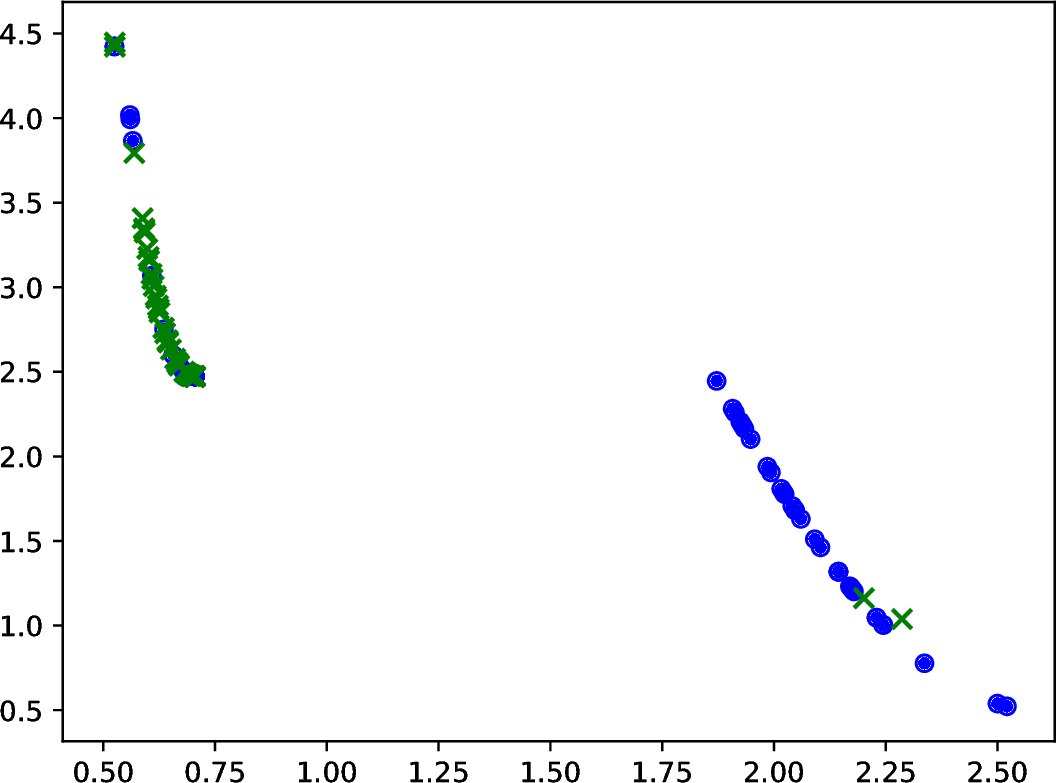}}
  \subfigure[WFG48 - URAW]{\includegraphics[width=0.245\linewidth]{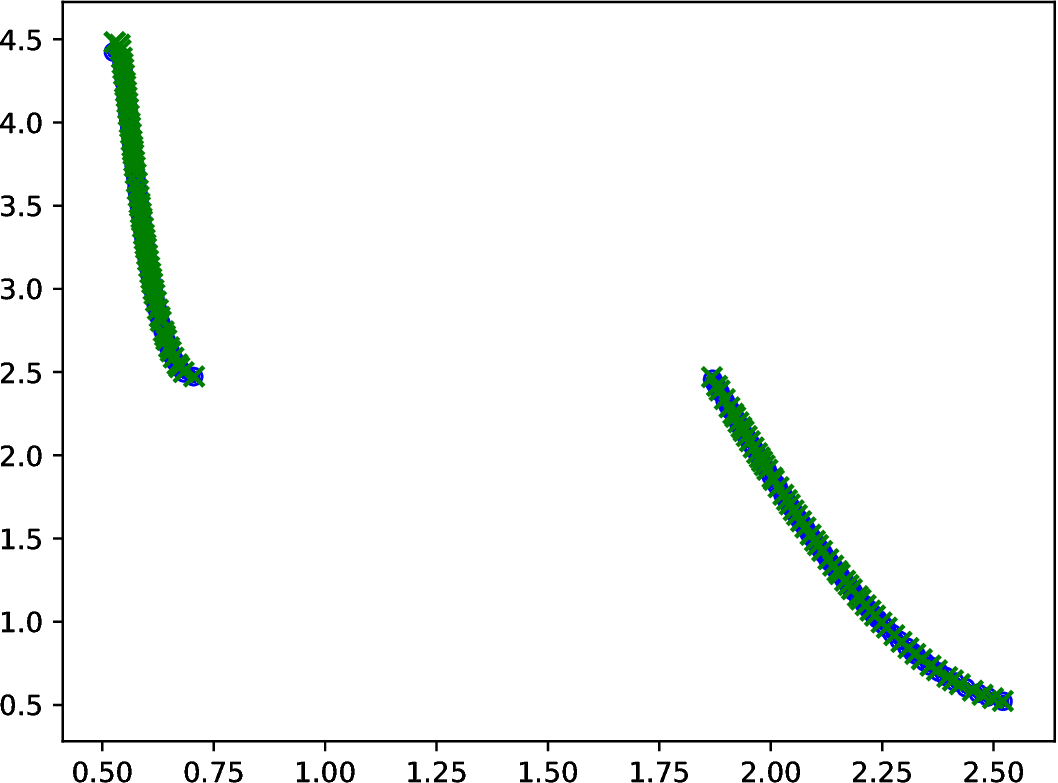}}
  
  \caption{Final solutions set with the best (blue circle) and median (green cross sign) HV metric values obtained by the algorithms on WFG45 to WFG48.}
  \label{fig:wfg45-48}
\end{figure*}

The final solution set with the median HV metric values for 2 objectives is shown in \Cref{fig:wfg41-44,fig:wfg45-48}. It can be seen that the PFs of MOEA/D-URAW are mostly more evenly distributed. The adaptation of the weight vectors in the UR initialization method allows that it happens.

Secondly, \tref{tab:maopResults} and \tref{tab:maopMedianBestResults} shows the HV results for MaOPs with 4, 5, and 6 objectives. Here, MOEA/D-URAW continues to present better result in most of the test instances. The problem with WFG43 persists for all the cases, where Das and Dennis, and Uniformly Randomly outperforms it.

Moreover, for 5 and 6 objectives, WFG44 appears as a problem for MOEA/D-URAW. For the WFG47, the MOEA/D-URAW is not the best in the 4 and 5 objectives, in these cases, its version without adaptation of weights, UR, presented better performance. 

These problems presented by MOEA/D-URAW to the WFG43 and WFG44 are interesting to analyze. The fact is that both of them have strong shapes, concave and convex, respectively. It may impact the effectiveness of MOEA/D-URAW. The WFG47 result problems for 2 objectives was unexpected, and it is a key point of future investigation.	

\section{Conclusions}
\label{sec:conclusions}

	This work introduces the MOEA/D-URAW, a model that uses UR initialization method combined with weights adaptation in order to obtain both flexible population size and better adapted final solution sets. Its performance was evaluated using MOOPs and MaOPs with different PF shapes.

The results indicate that the MOEA/D-URAW presents a better performance, specifically in WFG problems with concave, convex, mixed, linear, convex and disconnected hyperplane PF shapes. The proposed model is better than the other methodologies evaluated in 72.5\% of cases, considering all problems and objectives tested. However, it is important to note that the proposed adaptation did not significantly improve the results when the PF shape is disconnected and convex. In 22.5\% of the results, there were statistically significant differences. In these cases, the PFs have strong concave or strong convex formats. These results indicate that the use of sparsity level may not be appropriate to adapt the weights with these PFs formats. There is still a case in which the TSF approach performed better than the others, i.e., the WFG47 with 2 objectives. This case has a PF with disconnected and concave shape, and the reason why this happens will be investigated in more detail in a future work.

Another relevant issue is that, in the literature, there is no consensus on the best frequency of weight adaptation. Different frequencies affect the outcome of the approaches that use weight adaptation. As future work, it is worth investigating the impact of this parameter in approaches that use weights adaptation.

    \begin{table*}[!ht]
   \centering
   \caption{Average and standard deviation of HV results obtained in 100 independent runs on WFG41 to WFG48 on MOOP. The highest average is highlighted with gray background. $\star$ indicates whether the MOEA/D-URAW is significantly better than all other models or which algorithm presented significantly better results than it.}
   \begin{threeparttable}
   \begin{tabular}{c cccc | cccc}
   \toprule 
   \multirow{2}{*}{\textbf{Problems}} & \multicolumn{4}{c}{\textbf{2 Objectives}} & \multicolumn{4}{c}{\textbf{3 Objectives}}\\
   \cmidrule(r){2-5}
   \cmidrule(r){6-9}
    & \textbf{DD} & \textbf{UR} & \textbf{TSF} & \textbf{URAW} &
    \textbf{DD} & \textbf{UR} & \textbf{TSF} & \textbf{URAW}\\

\midrule																																		
	\multirow{2}{*}{\textbf{WFG41}}	&	0.6622	&	0.6567	&			0.6620			&	\cellcolor{gray!25}		0.6669		$\star$	&			1.0736			&	1.0672	&	0.9665	&	\cellcolor{gray!25}		1.1145		$\star$	\\
	\textbf{}	&	1.50E-03	&	4.80E-03	&			1.30E-02			&	\cellcolor{gray!25}		1.60E-03			&			2.70E-03			&	8.10E-03	&	2.90E-02	&	\cellcolor{gray!25}		3.30E-03			\\
\midrule																																		
	\multirow{2}{*}{\textbf{WFG42}}	&	1.2078	&	1.206	&			1.1994			&	\cellcolor{gray!25}		1.2095		$\star$	&			1.652			&	1.6829	&	1.6408	&	\cellcolor{gray!25}		1.6906		$\star$	\\
	\textbf{}	&	9.00E-04	&	1.20E-03	&			4.00E-03			&	\cellcolor{gray!25}		6.00E-04			&			6.50E-03			&	2.70E-03	&	9.00E-03	&	\cellcolor{gray!25}		1.70E-03			\\
\midrule																																		
	\multirow{2}{*}{\textbf{WFG43}}	&	0.4806	&	0.4798	&			0.4831			&	\cellcolor{gray!25}		0.5015			&	\cellcolor{gray!25}		0.6533		$\star$	&	0.6401	&	0.5088	&			0.6416			\\
	\textbf{}	&	5.10E-03	&	2.10E-03	&			5.60E-02			&	\cellcolor{gray!25}		3.70E-03			&	\cellcolor{gray!25}		1.70E-03			&	4.10E-03	&	4.70E-02	&			8.60E-03			\\
\midrule																																		
	\multirow{2}{*}{\textbf{WFG44}}	&	1.3865	&	1.3859	&			1.3802			&	\cellcolor{gray!25}		1.3909		$\star$	&			1.7225			&	1.7243	&	1.6887	&	\cellcolor{gray!25}		1.7260		$\star$	\\
	\textbf{}	&	1.20E-03	&	8.00E-04	&			3.70E-03			&	\cellcolor{gray!25}		5.00E-04			&			3.50E-03			&	1.20E-03	&	7.70E-03	&	\cellcolor{gray!25}		1.40E-03			\\
\midrule																																		
	\multirow{2}{*}{\textbf{WFG45}}	&	0.8091	&	0.8074	&			0.8098			&	\cellcolor{gray!25}		0.8109			&			1.1995			&	1.2181	&	1.1167	&	\cellcolor{gray!25}		1.2524		$\star$	\\
	\textbf{}	&	1.60E-03	&	2.90E-03	&			5.70E-03			&	\cellcolor{gray!25}		1.40E-03			&			5.20E-03			&	7.10E-03	&	2.70E-02	&	\cellcolor{gray!25}		3.50E-03			\\
\midrule																																		
	\multirow{2}{*}{\textbf{WFG46}}	&	0.9354	&	0.9278	&			0.9268			&	\cellcolor{gray!25}		0.9369		$\star$	&			1.4210			&	1.4546	&	1.3606	&	\cellcolor{gray!25}		1.4866		$\star$	\\
	\textbf{}	&	1.60E-03	&	3.90E-03	&			1.10E-02			&	\cellcolor{gray!25}		1.20E-03			&			5.20E-03			&	6.80E-03	&	2.00E-02	&	\cellcolor{gray!25}		2.80E-03			\\
\midrule																																		
	\multirow{2}{*}{\textbf{WFG47}}	&	0.8012	&	0.8014	&	\cellcolor{gray!25}		0.8474		$\star$	&			0.8179			&			1.2521			&	1.2648	&	1.1487	&	\cellcolor{gray!25}		1.2773		$\star$	\\
	\textbf{}	&	5.50E-02	&	5.30E-02	&	\cellcolor{gray!25}		5.30E-02			&			5.30E-02			&			1.40E-02			&	5.70E-03	&	2.80E-02	&	\cellcolor{gray!25}		3.00E-02			\\
\midrule																																		
	\multirow{2}{*}{\textbf{WFG48}}	&	0.9317	&	0.9509	&			0.9357			&	\cellcolor{gray!25}		0.9526		$\star$	&			1.6108			&	1.6212	&	1.4753	&	\cellcolor{gray!25}		1.6308		$\star$	\\
	\textbf{}	&	9.80E-02	&	9.50E-02	&			6.70E-02			&	\cellcolor{gray!25}		9.70E-02			&			9.00E-03			&	3.00E-03	&	4.30E-02	&	\cellcolor{gray!25}		2.20E-03			\\

   \bottomrule
   \end{tabular}
   \end{threeparttable}
   \label{tab:mopResults}
\end{table*}

	\begin{table*}[!ht]
   \centering
   \caption{The best and median HV metric results obtained in 100 independent runs on WFG41 to WFG48 on MOOP. The highest values in the results are highlighted in bold.} 
   \begin{threeparttable}
   \begin{tabular}{c cccc | cccc}
   \toprule 
   \multirow{2}{*}{\textbf{Problems}} & \multicolumn{4}{c}{\textbf{2 Objectives}} & \multicolumn{4}{c}{\textbf{3 Objectives}}\\
   \cmidrule(r){2-5}
   \cmidrule(r){6-9}
    & \textbf{DD} & \textbf{UR} & \textbf{TSF} & \textbf{URAW} &
    \textbf{DD} & \textbf{UR} & \textbf{TSF} & \textbf{URAW}\\

\midrule		
	\multirow{2}{*}{\textbf{WFG41}}	&	0.6635	&	0.6628	&	\textbf{	0.6768	}	&		0.6687		&		1.0788		&	1.0811	&	1.0238	&	\textbf{	1.1196	}	\\
	\textbf{}	&	0.6625	&	0.6580	&		0.6647		&	\textbf{	0.6675	}	&		1.0744		&	1.0684	&	0.9676	&	\textbf{	1.1155	}	\\
\midrule																										
	\multirow{2}{*}{\textbf{WFG42}}	&	1.2088	&	1.2080	&		1.2049		&	\textbf{	1.2102	}	&		1.6620		&	1.6871	&	1.6653	&	\textbf{	1.6924	}	\\
	\textbf{}	&	1.2079	&	1.2061	&		1.2004		&	\textbf{	1.2093	}	&		1.6530		&	1.6834	&	1.6404	&	\textbf{	1.6908	}	\\
\midrule																										
	\multirow{2}{*}{\textbf{WFG43}}	&	0.4829	&	0.4828	&	\textbf{	0.5417	}	&		0.5053		&	\textbf{	0.6561	}	&	0.6467	&	0.6373	&		0.6508		\\
	\textbf{}	&	0.4815	&	0.4802	&	\textbf{	0.5096	}	&		0.5030		&	\textbf{	0.6535	}	&	0.6410	&	0.5110	&		0.6434		\\
\midrule																										
	\multirow{2}{*}{\textbf{WFG44}}	&	1.3880	&	1.3871	&		1.3870		&	\textbf{	1.3914	}	&		1.7252		&	1.7264	&	1.7015	&	\textbf{	1.7272	}	\\
	\textbf{}	&	1.3866	&	1.3861	&		1.3810		&	\textbf{	1.3912	}	&		1.7228		&	1.7241	&	1.6907	&	\textbf{	1.7268	}	\\
\midrule																										
	\multirow{2}{*}{\textbf{WFG45}}	&	0.8104	&	0.8116	&	\textbf{	0.8173	}	&		0.8126		&		1.2063		&	1.2298	&	1.1732	&	\textbf{	1.2586	}	\\
	\textbf{}	&	0.8095	&	0.8081	&	\textbf{	0.8115	}	&		0.8111		&		1.2006		&	1.2196	&	1.1194	&	\textbf{	1.2529	}	\\
\midrule																										
	\multirow{2}{*}{\textbf{WFG46}}	&	0.9368	&	0.9317	&	\textbf{	0.9388	}	&		0.9382		&		1.4308		&	1.4665	&	1.4006	&	\textbf{	1.4908	}	\\
	\textbf{}	&	0.9358	&	0.9288	&		0.9302		&	\textbf{	0.9371	}	&		1.4227		&	1.4559	&	1.3616	&	\textbf{	1.4871	}	\\
\midrule																										
	\multirow{2}{*}{\textbf{WFG47}}	&	0.8621	&	0.8440	&	\textbf{	0.9104	}	&		0.8557		&		1.2689		&	1.2759	&	1.2186	&	\textbf{	1.2856	}	\\
	\textbf{}	&	0.7511	&	0.8414	&	\textbf{	0.8643	}	&		0.8545		&		1.2550		&	1.2660	&	1.1515	&	\textbf{	1.2807	}	\\
\midrule																										
	\multirow{2}{*}{\textbf{WFG48}}	&	1.0320	&	1.0327	&		1.0246		&	\textbf{	1.0337	}	&		1.6225		&	1.6267	&	1.5558	&	\textbf{	1.6344	}	\\
	\textbf{}	&	0.8377	&	1.0290	&		0.9647		&	\textbf{	1.0324	}	&		1.6137		&	1.6218	&	1.4806	&	\textbf{	1.6314	}	\\

   \bottomrule
   \end{tabular}
   \end{threeparttable}
   \label{tab:mopMedianBestResults}
\end{table*}

    \begin{table*}[!ht]
   \centering
   \caption{Average and standard deviation of HV results obtained in 100 independent runs on WFG41 to WFG48 on MaOP. The highest average is highlighted with gray background. $\star$ indicates whether the MOEA/D-URAW is significantly better than all other models or which algorithm presented significantly better results than it.}
   \begin{threeparttable}
   \begin{tabular}{c cccc | cccc | cccc}
   \toprule 
   \multirow{2}{*}{\textbf{Problems}} & \multicolumn{4}{c}{\textbf{4 Objectives}} & \multicolumn{4}{c}{\textbf{5 Objectives}} & \multicolumn{4}{c}{\textbf{6 Objectives}}\\
   \cmidrule(r){2-5}
   \cmidrule(r){6-9}
   \cmidrule(r){10-13}
    & \textbf{DD} & \textbf{UR} & \textbf{TSF} & \textbf{URAW} &
    \textbf{DD} & \textbf{UR} & \textbf{TSF} & \textbf{URAW} &
    \textbf{DD} & \textbf{UR} & \textbf{TSF} & \textbf{URAW}\\

\midrule																																																						
	\multirow{2}{*}{\textbf{WFG41}}	&	1.340	&			1.464			&	1.145	&	\cellcolor{gray!25}		1.494		$\star$	&			1.563			&			1.840			&	1.312	&	\cellcolor{gray!25}		1.867		$\star$	&			1.812			&	2.215	&	1.513	&	\cellcolor{gray!25}		2.261		$\star$	\\
	\textbf{}	&	2.6E-02	&			1.1E-02			&	4.1E-02	&	\cellcolor{gray!25}		8.4E-03			&			1.6E-02			&			1.5E-02			&	4.6E-02	&	\cellcolor{gray!25}		1.3E-02			&			4.2E-02			&	1.9E-02	&	5.1E-02	&	\cellcolor{gray!25}		2.4E-02			\\
\midrule																																																						
	\multirow{2}{*}{\textbf{WFG42}}	&	2.011	&			2.046			&	1.942	&	\cellcolor{gray!25}		2.056		$\star$	&			2.451			&			2.460			&	2.300	&	\cellcolor{gray!25}		2.471		$\star$	&			2.939			&	2.939	&	2.701	&	\cellcolor{gray!25}		2.961		$\star$	\\
	\textbf{}	&	1.5E-02	&			5.9E-03			&	1.5E-02	&	\cellcolor{gray!25}		4.1E-03			&			5.3E-03			&			1.1E-02			&	3.0E-02	&	\cellcolor{gray!25}		7.0E-03			&			2.1E-02			&	1.9E-02	&	5.3E-02	&	\cellcolor{gray!25}		1.3E-02			\\
\midrule																																																						
	\multirow{2}{*}{\textbf{WFG43}}	&	0.821	&	\cellcolor{gray!25}		0.832		$\star$	&	0.521	&			0.819			&			1.009			&	\cellcolor{gray!25}		1.032		$\star$	&	0.537	&			0.998			&	\cellcolor{gray!25}		1.248		$\star$	&	1.242	&	0.582	&			1.144			\\
	\textbf{}	&	4.9E-03	&	\cellcolor{gray!25}		5.4E-03			&	5.1E-02	&			1.7E-02			&			1.0E-02			&	\cellcolor{gray!25}		8.9E-03			&	5.1E-02	&			2.6E-02			&	\cellcolor{gray!25}		1.4E-02			&	1.9E-02	&	6.1E-02	&			5.8E-02			\\
\midrule																																																						
	\multirow{2}{*}{\textbf{WFG44}}	&	2.067	&			2.060			&	1.901	&	\cellcolor{gray!25}		2.068		$\star$	&	\cellcolor{gray!25}		2.481		$\star$	&			2.458			&	2.070	&			2.472			&	\cellcolor{gray!25}		2.964		$\star$	&	2.861	&	2.165	&			2.871			\\
	\textbf{}	&	5.3E-03	&			9.8E-03			&	3.6E-02	&	\cellcolor{gray!25}		5.3E-03			&	\cellcolor{gray!25}		9.2E-03			&			3.2E-02			&	1.4E-01	&			1.4E-02			&	\cellcolor{gray!25}		3.1E-02			&	1.6E-01	&	4.8E-01	&			1.5E-01			\\
\midrule																																																						
	\multirow{2}{*}{\textbf{WFG45}}	&	1.428	&			1.602			&	1.331	&	\cellcolor{gray!25}		1.628		$\star$	&			1.617			&			1.984			&	1.553	&	\cellcolor{gray!25}		2.002		$\star$	&			1.860			&	2.375	&	1.818	&	\cellcolor{gray!25}		2.400		$\star$	\\
	\textbf{}	&	2.2E-02	&			1.1E-02			&	3.4E-02	&	\cellcolor{gray!25}		8.7E-03			&			2.7E-02			&			1.5E-02			&	3.7E-02	&	\cellcolor{gray!25}		1.2E-02			&			3.3E-02			&	2.5E-02	&	5.0E-02	&	\cellcolor{gray!25}		2.3E-02			\\
\midrule																																																						
	\multirow{2}{*}{\textbf{WFG46}}	&	1.717	&			1.877			&	1.598	&	\cellcolor{gray!25}		1.904		$\star$	&			2.011			&			2.30			&	1.901	&	\cellcolor{gray!25}		2.337		$\star$	&			2.378			&	2.762	&	2.210	&	\cellcolor{gray!25}		2.820		$\star$	\\
	\textbf{}	&	1.7E-02	&			7.0E-03			&	3.0E-02	&	\cellcolor{gray!25}		6.5E-03			&			1.6E-02			&			1.2E-02			&	3.6E-02	&	\cellcolor{gray!25}		8.9E-03			&			4.6E-02			&	2.3E-02	&	4.1E-02	&	\cellcolor{gray!25}		1.6E-02			\\
\midrule																																																						
	\multirow{2}{*}{\textbf{WFG47}}	&	1.506	&	\cellcolor{gray!25}		1.650		$\star$	&	1.366	&			1.643			&			1.646			&	\cellcolor{gray!25}		2.029		$\star$	&	1.610	&			2.025			&			1.864			&	2.429	&	1.874	&	\cellcolor{gray!25}		2.435		$\star$	\\
	\textbf{}	&	7.3E-02	&	\cellcolor{gray!25}		8.1E-03			&	3.9E-02	&			6.6E-02			&			1.1E-01			&	\cellcolor{gray!25}		8.9E-02			&	3.7E-02	&			6.1E-02			&			8.0E-02			&	1.3E-01	&	5.8E-02	&	\cellcolor{gray!25}		7.5E-02			\\
\midrule																																																						
	\multirow{2}{*}{\textbf{WFG48}}	&	1.977	&			2.032			&	1.759	&	\cellcolor{gray!25}		2.036		$\star$	&			2.410			&			2.445			&	2.091	&	\cellcolor{gray!25}		2.457		$\star$	&			2.903			&	2.925	&	2.526	&	\cellcolor{gray!25}		2.944		$\star$	\\
	\textbf{}	&	1.7E-02	&			6.4E-03			&	6.3E-02	&	\cellcolor{gray!25}		8.1E-02			&			1.3E-02			&			1.2E-02			&	1.0E-01	&	\cellcolor{gray!25}		1.2E-02			&			2.9E-02			&	2.1E-02	&	1.1E-01	&	\cellcolor{gray!25}		1.8E-02			\\

   \bottomrule
   \end{tabular}
   \end{threeparttable}
   \label{tab:maopResults}
\end{table*}
	\begin{table*}[!ht]
   \centering
   \caption{The best and median HV metric results obtained in 100 independent runs on WFG41 to WFG48 on MaOP. The highest values in the results are highlighted in bold.}
   \begin{threeparttable}
   \begin{tabular}{c cccc | cccc | cccc}
   \toprule 
   \multirow{2}{*}{\textbf{Problems}} & \multicolumn{4}{c}{\textbf{4 Objectives}} & \multicolumn{4}{c}{\textbf{5 Objectives}} & \multicolumn{4}{c}{\textbf{6 Objectives}}\\
   \cmidrule(r){2-5}
   \cmidrule(r){6-9}
   \cmidrule(r){10-13}
    & \textbf{DD} & \textbf{UR} & \textbf{TSF} & \textbf{URAW} &
    \textbf{DD} & \textbf{UR} & \textbf{TSF} & \textbf{URAW} &
    \textbf{DD} & \textbf{UR} & \textbf{TSF} & \textbf{URAW}\\

\midrule		
	\multirow{2}{*}{\textbf{WFG41}}	&	1.3632	&		1.4878		&	1.2339	&	\textbf{	1.5110	}	&		1.5854		&		1.8730		&	1.4196	&	\textbf{	1.8906	}	&		1.8822		&		2.2670		&	1.6321	&	\textbf{	2.3094	}	\\
	\textbf{}	&	1.3480	&		1.4657		&	1.1504	&	\textbf{	1.4948	}	&		1.5656		&		1.8395		&	1.3206	&	\textbf{	1.8693	}	&		1.8264		&		2.2158		&	1.5104	&	\textbf{	2.2619	}	\\
\midrule																																										
	\multirow{2}{*}{\textbf{WFG42}}	&	2.0306	&		2.0557		&	1.9716	&	\textbf{	2.0619	}	&		2.4627		&		2.4717		&	2.3623	&	\textbf{	2.4812	}	&		2.9570		&		2.9657		&	2.7988	&	\textbf{	2.9769	}	\\
	\textbf{}	&	2.0102	&		2.0466		&	1.9441	&	\textbf{	2.0576	}	&		2.4519		&		2.4621		&	2.3021	&	\textbf{	2.4739	}	&		2.9451		&		2.9449		&	2.7096	&	\textbf{	2.9668	}	\\
\midrule																																										
	\multirow{2}{*}{\textbf{WFG43}}	&	0.8327	&		0.8405		&	0.6353	&	\textbf{	0.8432	}	&		1.0303		&	\textbf{	1.0513	}	&	0.6875	&		1.0383		&	\textbf{	1.2750	}	&		1.2731		&	0.7043	&		1.2245		\\
	\textbf{}	&	0.8212	&	\textbf{	0.8326	}	&	0.5283	&		0.8247		&		1.0105		&	\textbf{	1.0328	}	&	0.5343	&		1.0006		&	\textbf{	1.2498	}	&		1.2463		&	0.5926	&		1.1640		\\
\midrule																																										
	\multirow{2}{*}{\textbf{WFG44}}	&	2.0731	&		2.0718		&	1.9819	&	\textbf{	2.0734	}	&	\textbf{	2.4882	}	&		2.4869		&	2.3018	&		2.4879		&	\textbf{	2.9860	}	&		2.9833		&	2.7231	&		2.9841		\\
	\textbf{}	&	2.0678	&		2.0620		&	1.9032	&	\textbf{	2.0685	}	&	\textbf{	2.4786	}	&		2.4642		&	2.0749	&		2.4731		&	\textbf{	2.9694	}	&		2.9079		&	2.2491	&		2.9008		\\
\midrule																																										
	\multirow{2}{*}{\textbf{WFG45}}	&	1.4753	&		1.6228		&	1.4270	&	\textbf{	1.6428	}	&		1.6613		&		2.0136		&	1.6324	&	\textbf{	2.0289	}	&		1.9064		&		2.4430		&	1.9181	&	\textbf{	2.4475	}	\\
	\textbf{}	&	1.4314	&		1.6031		&	1.3342	&	\textbf{	1.6284	}	&		1.6235		&		1.9857		&	1.5530	&	\textbf{	2.0032	}	&		1.8678		&		2.3753		&	1.8242	&	\textbf{	2.4010	}	\\
\midrule																																										
	\multirow{2}{*}{\textbf{WFG46}}	&	1.7642	&		1.8910		&	1.6657	&	\textbf{	1.9155	}	&		2.0698		&		2.3211		&	1.9831	&	\textbf{	2.3536	}	&		2.4870		&		2.8109		&	2.3071	&	\textbf{	2.8461	}	\\
	\textbf{}	&	1.7148	&		1.8784		&	1.5997	&	\textbf{	1.9056	}	&		2.0135		&		2.3015		&	1.9044	&	\textbf{	2.3377	}	&		2.3920		&		2.7661		&	2.2160	&	\textbf{	2.8225	}	\\
\midrule																																										
	\multirow{2}{*}{\textbf{WFG47}}	&	1.6036	&		1.6656		&	1.4428	&	\textbf{	1.6698	}	&		1.8537		&	\textbf{	2.0642	}	&	1.6999	&		2.0552		&		2.0029		&	\textbf{	2.4939	}	&	2.0005	&		2.4918		\\
	\textbf{}	&	1.5072	&		1.6500		&	1.3639	&	\textbf{	1.6535	}	&		1.6232		&	\textbf{	2.0420	}	&	1.6145	&		2.0328		&		1.8890		&	\textbf{	2.4536	}	&	1.8813	&		2.4453		\\
\midrule																																										
	\multirow{2}{*}{\textbf{WFG48}}	&	2.0005	&		2.0406		&	1.8416	&	\textbf{	2.0427	}	&		2.4268		&		2.4640		&	2.2423	&	\textbf{	2.4679	}	&		2.9349		&		2.9608		&	2.6785	&	\textbf{	2.9668	}	\\
	\textbf{}	&	1.9786	&		2.0337		&	1.7771	&	\textbf{	2.0367	}	&		2.4157		&		2.4489		&	2.1112	&	\textbf{	2.4604	}	&		2.9120		&		2.9309		&	2.5407	&	\textbf{	2.9518	}	\\

   \bottomrule
   \end{tabular}
   \end{threeparttable}
   \label{tab:maopMedianBestResults}
\end{table*}


\begin{acks}
The authors would like to thank the Brazilian National Council for Technological and Scientific Development (CNPq) for partially financing this research study.
\end{acks}

\bibliographystyle{ACM-Reference-Format}
\bibliography{references} 

\end{document}